\documentclass[a4paper]{article}

\usepackage[american]{babel}
\usepackage[utf8]{inputenc}
\usepackage[T1]{fontenc}
\usepackage{lmodern}
\usepackage{csquotes}
\usepackage{fullpage}
\usepackage{authblk}
\usepackage{amsthm}
\usepackage{amssymb}
\usepackage{mathtools}
\usepackage{algorithm}
\usepackage{algorithmic}
\usepackage{pdflscape} 
\usepackage{tikz}
\usepackage[colorinlistoftodos,
            prependcaption,
            textsize=small]{todonotes}
\usepackage{microtype}
\usepackage[style = authoryear,
            maxbibnames = 100,
            maxcitenames = 2,
            isbn = false,
            backend=bibtex8]{biblatex}
\usepackage[colorlinks,
            citecolor=blue,
            urlcolor=blue,
            linkcolor=blue]{hyperref}
\vfuzz \hfuzz
\providecommand{\keywords}[1]{\noindent\textbf{Keywords:} #1}

\usetikzlibrary{external,arrows.meta,positioning}

\tikzsetexternalprefix{figures/}

\tikzexternaldisable

\newtheoremstyle{plain} 
{1em}
{1em}
{}
{}
{\bfseries}
{.}
{.5em}
{}

\bibliography{paper.bib}
\newcommand{\Prob}{\mathbf{Pr}}
\newcommand{\argmax}{\mbox{arg}\max}
\newcommand{\argmin}{\mbox{arg}\min}

 \usepackage{subcaption}
 \usepackage{algorithm}
 \usepackage{enumitem}
 \usepackage{tabularx}
 \usepackage{booktabs}
 \usepackage{xcolor}
 \usepackage{stmaryrd} 

\begin{document}

\title{
Learn-n-Route: Learning implicit preferences for vehicle routing
}

\author[1]{Rocsildes Canoy}
\author[1]{V\'ictor Bucarey}
\author[1]{Jayanta Mandi}
\author[1]{Tias Guns}
\affil[1]{Vrije Universiteit Brussel, Brussels, Belgium.}
\affil[ ]{ \textit{\{firstname.lastname\}@vub.be}}

\maketitle

\begin{abstract}
We investigate a \textit{learning} decision support system for vehicle routing, where the routing engine learns implicit preferences that human planners have when manually creating route plans (or \textit{routings}). The goal is to use these learned \textit{subjective} preferences on top of the distance-based \textit{objective} criterion in vehicle routing systems. This is an alternative to the practice of distinctively formulating a custom VRP for every company with its own routing requirements. Instead, we assume the presence of past vehicle routing solutions over similar sets of customers, and learn to make similar choices. The learning approach is based on the concept of learning a
Markov model, which corresponds to a probabilistic transition matrix, rather than a deterministic distance matrix. This nevertheless allows us to use existing arc routing VRP software in creating the actual routings, and to optimize over both distances and preferences at the same time. For the learning, we explore different schemes to construct the probabilistic transition matrix that can co-evolve with changing preferences over time. Our results 
on a use-case with a small transportation company show that our method is able to generate results that are close to the manually created solutions, without needing to characterize all constraints and sub-objectives explicitly. Even in the case of changes in the customer sets, our method is able to find solutions that are closer to the actual routings than when using only distances, and hence, solutions that require fewer manual changes when transformed into practical routings. 
\\

  \keywords{vehicle routing, preference learning, Markov models}
\end{abstract}


\section{Introduction}

Vehicle routing problems (VRP) at small or medium-sized enterprises (SME)  is constrained by the limited number of vehicles, the capacity of each delivery vehicle, and the scheduling horizon within which all deliveries have to be made. The objective, often implicitly, can include a wide range of company goals including reducing operational costs, minimizing fuel consumption and carbon emissions, as well as optimizing driver familiarity with the routes and maximizing fairness by assigning tours of similar length and/or time duration to the drivers. Daily plans are often created in a route optimization software that is capable of producing plans that are optimal in terms of route length and travel time. We have observed, however, that in practice, route planners usually modify the result given by the software, or simply pull out, modify, and reuse an old plan that has been used and known to work in the past. The planners, by performing these modifications, are essentially optimizing with their own set of objectives and personal preferences.

These preferences are often subjective and sometimes delicate to formalize in constraints. Some examples are: where the best places for lunch breaks are, which stops are best served earlier or later, and which drivers (tours) are more flexible in receiving more stops. Failure to capture such \textit{human} aspects is often a source of frustration for both drivers and planners, and a cause for reluctance to use the optimisation software. Furthermore, planners may find it easier or more effective to manually change a previous solution than to provide or update the detailed information in the system. Being able to automatically capture such preferences without the need to formalize them can hence lead to a wider acceptance and better use of optimisation systems.

The goal of this research is to learn the preferences of the planners (and implicitly the drivers) when choosing one option over another. 
Hence, the goal is to build an `intelligent' system that can effectively reuse all of the knowledge and effort that have been put into creating previous routings; much like how human planners use prior experience. We focus on techniques from the domain of artificial intelligence that can learn from historical data, and that can be used to manage and recommend similar routes as used in the past. This is in contrast to the current practice of optimizing a separate VRP instance each day. 


To learn from historical data, we take inspiration from various machine learning papers on route prediction for a single vehicle: Markov models developed from historical data have been applied to driver turn prediction [\cite{krumm2016markov}], prediction of the remainder of the route by looking at the previous road segments taken by the driver [\cite{ashbrook2003using}], and predicting individual road choices given the origin and destination [\cite{simmons2006learning}].
These studies have produced positive and encouraging results for their respective tasks. Hence, in this work, we investigate the use of Markov models for predicting the route choices for an entire fleet, and how to use these choices to create {\it preference-aware} solutions to the VRP.

With a 
Markov model, route optimization can be done by maximizing the product of the probabilities of the sequential actions taken by the vehicles, which corresponds to maximizing the sum of log likelihoods over the arcs. In the case of a first order Markov model, a key property of our approach is that it can reuse \textit{any existing VRP solution method} to find the maximum likelihood solution, as it corresponds to the sum of log likelihoods over the individual arcs. This is a promising novel approach to the vehicle routing problem.

Our proposed first order Markov model approach has been published in a conference proceeding [\cite{canoy2019vehicle}] and in this article, we extend this methodology by developing the following new contributions:

\begin{itemize}
    \item We provide a more general formalization of the methods for learning and optimizing the preferences,  and detail both a first order \emph{and a second order} Markov model.
    \item We introduce a new and superior weighing technique called the exponential weighing scheme and examine how it handles data drift, i.e., when there is either a large reduction or a sudden increase in the number of stops. 
    \item We present an extended study on distance-based, preference-based, and combined optimisation.
\end{itemize}

This manuscript is organized as follows: in Section \ref{s:related}, we provide the relevant work related to this article. We discuss in Section \ref{s:formalisation} routing models that maximize the probability of the learned preferences. Later in the section, we introduce the algorithms to learn the transition matrix from historical data. The comparison of the different construction schemes and the experimental results on actual company data are shown in Section \ref{s:expts}. Finally, our conclusions and future research directions are presented in Section \ref{s:conclusions}.

\section{Related Work} \label{s:related}

The first mathematical formulation and algorithmic approach to solving the Vehicle Routing Problem (VRP) appeared in the paper by  \cite{dantzig1959truck} which aimed to find an optimal routing for a fleet of gasoline delivery trucks. Since its introduction, the VRP has become one of the most studied combinatorial optimization problems. Faced on a daily basis by distributors and logistics companies worldwide, the problem has attracted a lot of attention due to its significant economic importance.

A large part of the research effort concerning the VRP has focused on its classical and basic version---the \emph{Capacitated} Vehicle Routing Problem (CVRP). The presumption is that the algorithms developed for CVRP could be extended and applied to more complicated real-world cases [\cite{laporte2007you}]. 
Due to the recent development of new and more efficient optimisation methods, research interest has shifted towards realistic VRP variants known as Rich VRP [\cite{caceres2015rich,drexl2012rich}]. These problems deal with realistic, and sometimes multi-ob\-jec\-tive, optimisation functions, uncertainty, and a wide variety of real-life constraints related to time and distance factors, inventory and scheduling, environmental and energy concerns, driver experience, etc. [\cite{ mor2020vehicle}]. 

The VRP becomes increasingly complex as additional sub-ob\-jec\-tives and constraints are introduced. The inclusion of preferences, for example, necessitates the difficult, if not impossible, task of formalizing the route planners' knowledge and choice preferences explicitly in terms of constraints and weights. In most cases, it is much easier to get examples and historical solutions rather than to extract explicit decision rules from the planners, as observed by Potvin et al. in the case of vehicle dispatching [\cite{potvin1993learning}]. One approach is to use learning techniques, particularly learning by examples, to reproduce the planners' decision behavior. To this end, we develop a new method that learns from previous solutions by using a Markov model, which can simplify the problem formulation phase of vehicle routing by eliminating the need to characterize all preference constraints and sub-ob\-jec\-tives explicitly.

Learning from historical solutions has been investigated before within the context of constraint programming. For example, in the paper of Beldiceanu and Simonis on constraint seeker [\cite{beldiceanu2011constraint}] and model seeker [\cite{beldiceanu2012model}], and Picard-Cantin et al. on learning constraint parameters from data [\cite{picard2016learning}], where a Markov chain is used, but for individual constraints. In this respect, our goal is not to learn constraint instantiations, but rather to learn choice preferences, e.g., as part of the objective. 
Related to the latter is the work on Constructive Preference Elicitation~[\cite{dragone2018constructive}], although this elicitation technique actively queries the user, as does constraint acquisition~[\cite{bessiere2017constraint}]. 

Our motivation for Markov models is that they have been previously used in route prediction of individual vehicles. \cite{ashbrook2003using} presented a Markov model to predict the next stop of a driving vehicle. They estimated transition probabilities as frequencies and considered Markov models that depend not only on the actual stop but also on past stops; that is, high order Markov chains.  Personalized route prediction has been used in transportation systems that provide drivers with real-time traffic information and in intelligent vehicle systems for optimizing energy efficiency in hybrid vehicles~[\cite{deguchi2004hev}].  \cite{ye2015method} introduced a route prediction method that can accurately predict an entire route early in the trip. The method is based on Hidden Markov Models (HMM) trained from the driver's past history. A route prediction algorithm that predicts a driving route for a given pair of origin and destination was presented by \cite{wang2015building}. Also based on the first order Markov model, the algorithm uses a probability transition matrix that was constructed to represent the knowledge of the driver's preferred links and routes.  An algorithm for driver turn prediction using a Markov model is developed in \cite{krumm2016markov}. Trained from the driver's historical data, the model makes a probabilistic prediction based on a short sequence of just-driven road segments. Experimental results showed that by looking at the most recent 10 segments into the past, the model can effectively predict the next segment with about 90\% accuracy. Another learning approach for driving preferences is to learn a linear cost function depending on several attributes, e.g., time, distance, fuel consumption, number of edges in the route, etc., as presented in \cite{potvin1993learning, funke2016deducing}, where linear programs to compute the weights were proposed. While related to our current work, we do not assume that a partial route is already given. Moreover, in our case, we are interested in optimizing the routing across \textit{multiple vehicles} rather than making predictions for a single vehicle.

Several algorithms were deployed to direct users through paths between origin and destination. This deployment is done either through an enumeration of paths [\cite{yang2015toward}], or by adapting weights on edges and computing shortest paths [\cite{letchner2006trip, delling2015navigation, guo2020context}]. Although similar to this second category, where weights represent probabilities, our work differs in that instead of planning single paths (i.e., for an origin-destination pair), our methods use preferences to create a routing plan for several vehicles. Also, we are adapting the classical CVRP mixed integer formulation in order to generate high probability routes.

The importance of preference-based learning can be realized from the study of \cite{ceikute2013routing}, which finds that local drivers often prefer paths that are not optimal in terms of travel time or cost. This is because local drivers favor routes based on their personal preferences. 
\cite{chang2011discovering} and \cite{letchner2006trip} proposed a framework to recommend personalized routes to the driver by taking into account which roads he or she is familiar with. Yang et al. estimated a driver's preferences by comparing the paths used by the driver to the skyline paths~[\cite{yang2014stochastic}]. \cite{yang2019pathrank} introduced \emph{PathRank}, a machine learning model trained with historical data, to rank candidate paths based on the driver's preferences. Once again, the above-mentioned are mostly concerned with point-to-point `shortest path' travel where the source and destination pairs are known. Our work, on the other hand, focuses on VRP: recommending a multi-vehicle solution over shared stops. 
Finally, a recent article showed an application of routing with preferences using \textit{patient preferences} over the drivers/vehicles to create route visits of caregivers, where the problem is formulated as a bi-criteria optimization problem [\cite{ait2019bicriteria}]. In contrast, our work is concerned in creating routings while taking into account the implicit \textit{planner and driver preferences} over the set of customers.


\section{Formalisation} \label{s:formalisation}

In this section we introduce the probabilistic model underlying our methodology to learn the drivers' preferences. Furthermore, we adapt the classical formulation of the CVRP to compute the route with the highest probability. 

\subsection{Routing, probabilities and Markov models}

We begin with a set of stops $V= \{0,1,\ldots,n\}$, where 0 represents the depot, and the other stops represent the locations of the customers, and a  fully-connected directed graph $G=(V,A)$. 
We call a \textbf{routing} $\mathbf{x}$ of $V$ with $m$ homogeneous vehicles, a set of at most $m$ tours in $G$ with each tour starting from and ending at the depot $0$ and such that each node in $V\backslash \{0\}$ is visited exactly once. 

We denote the set of all possible routings of $m$ vehicles over $V$ as ${\mathcal X}_V^m$, or simply ${\mathcal X}$ when it is clear from the context. 
Our goal is to determine a probabilistic model $\Prob$ such that $\Prob(\mathbf{x})$ is the probability of the vehicles following the routing $\mathbf{x}\in \mathcal{X}$.

A Markov model is defined over a single sequence of events, and so to be able to use Markov models for $\Prob(\mathbf{x}),$ we first daisy-chain the set of routes $\mathbf{x}.$ This is done by connecting the routes into one long sequence: 
\begin{align*}
&0\ \longrightarrow\ s_{11}\ \longrightarrow\ s_{12}\ \longrightarrow\ \ldots\ \longrightarrow\ s_{1p-1}\ \longrightarrow\ s_{1p}\ \longrightarrow\ 0\ \longrightarrow \\
& 0\ \longrightarrow\ s_{21}\ \longrightarrow\ s_{22}\ \longrightarrow\ \ldots\ \longrightarrow\ s_{2q-1}\ \longrightarrow\ s_{2q}\ \longrightarrow\ 0\ \longrightarrow \\
& \vdots \\
& 0\ \longrightarrow\ s_{m1}\ \longrightarrow\ s_{m2}\ \longrightarrow\ \ldots\ \longrightarrow\ s_{mr-1}\ \longrightarrow\ s_{mr}\ \longrightarrow\ 0, 
\end{align*}
which we then simplify by replacing the $(0\rightarrow0)$ connections by $0.$
Given this sequence interpretation of $\mathbf{x}$, we can decompose the joint probability $\Prob(\mathbf{x})$ as the probability of each element conditional on the elements before it in the sequence:
\begin{align*}
    \Prob(\mathbf{x})\ =&\  \Prob( 0,s_{11},s_{12},s_{13},\ldots,0,s_{21},\ldots) \\ =&\ \Prob(X_0 = 0)\,\Prob(X_1 = s_{11}\,|\,X_0 = 0)\,\Prob(X_2 = s_{12}\,|\,X_0 = 0,X_1 = s_{11}) \\
    &\ \qquad \Prob(X_3 = s_{13}\,|\,X_0 =0,X_1 = s_{11},X_2 = s_{12})\cdots.
\end{align*}
Estimating all conditional probabilities $\Prob(s_{ij}\,|\,0,\ldots,s_{ij-2},s_{ij-1})$ is difficult because of the large number of such possible probabilities and their sparsity in the data.
A common approach 
is to use a Markovian approximation  of the probabilistic model [\cite{ames1989markov}], which states that the probability of an event depends only on the state of the previous event. 
The main advantage is that, with this approach, the model can be described with much fewer parameters to learn.

We consider a \textit{first order} Markov chain $\{\bf{ X}_t\}_{t\ge 0}$ over $V$ with transition probabilities between states as:
\begin{align*}
\Prob& \left(X_{t+1} = j \, |\,  X_t = i, X_{t-1} = i_{t-1}, \ldots, X_0 = i_0 \right) \ \approx \ \mathbf{Pr}\left(X_{t+1} = j\,|\,X_t = i \right).
\end{align*}
The joint probability therefore becomes:
\begin{align*}
    \Prob(\mathbf{x})\ \approx\ \Prob(X_0= 0)\,\Prob(X_1 = s_{11}\,|\,X_0= 0)\,\Prob(X_2 = s_{12}\,|\,X_1 = s_{11})\,\Prob(X_3 = s_{13}\,|\,X_2 = s_{12})\cdots.
\end{align*}
On one hand, the interpretation of this model's assumption is straightforward: a driver's decision to go from a stop $i$ to another stop $j$ depends only on the current position (current stop), and does not consider the stops visited before that. With this assumption, the Markov model can be seen as a probability distribution over the set of arcs $A$.  On the other hand, this model is in fact an approximation: in practice, drivers and planners also take the stops preceding the current stop, and potentially even the stops after it, into account when making their decision. In either case, the current stop $X_t$ is the key deciding factor.

A finer-grained approximation is to consider a higher order, in particular a {\it second order}, Markov chain. In this case, the state of an event contains both the current stop and the stop before it. By extending the history of previously visited arcs, we get a better approximation of the probabilistic model over $\mathcal{X}.$ For the second order model, the conditional probabilities are approximated as follows:
\begin{align*}
\mathbf{Pr}& \left(X_{t+1} = k \, |\,  X_t = j, X_{t-1} = i, \ldots, X_0 = i_0 \right) \ \approx \ \mathbf{Pr}\left(X_{t+1} = k \, |\, X_{t - 1} = i,\, X_t = j \right).
\end{align*}
The trade-off is that the number of parameters to estimate increases from $O(n^2)$ to $O(n^3)$.  
Later in this section, we will investigate the different ways of estimating the conditional probabilities from the historical routings.

\subsection{Maximum likelihood routing} \label{sec:mlrouting}
Given a (learned) probability distribution $\mathbf{Pr}$ over a set of stops $V$, the goal is to find the \textit{maximum likelihood} routing, that is, the routing with the highest joint probability. Let ${\mathcal X}$ be the set of all possible routings for a given number $m$ of vehicles and a set of stops $V$.
The goal of finding the maximum likelihood routing then becomes:
\begin{equation}
    \max_{\mathbf{x}\in \mathcal{X} }\ \mathbf{Pr}(\mathbf{x}). \label{eq:ml_pr}
\end{equation}
Naturally, the set of all possible routings $\mathcal{X}$ is not given explicitly. However, we can define it implicitly as a set of constraints over decision variables, as common in optimisation approaches to vehicle routing. We first look into formulating the constraints, and then the objective function corresponding to Eq.~\eqref{eq:ml_pr}.

\subsubsection*{Constraints.}
Note that while the probability distribution $\mathbf{Pr}$ is defined over all possible routings, we are free to impose additional constraints over $\mathcal{X}$ when searching for the maximum likelihood routing. In effect, this will be a constrained maximum likelihood inference problem.

In this paper, we will use a standard Capacitated Vehicle Routing Problem (CVRP) formulation by means of Mixed Integer Programming (MIP) [\cite{toth2014family}]. To do so, we represent the routing $\mathcal{X}$ by the binary vector $\mathbf{x}\in \{0,1\}^{\vert A \vert}$ where $A$ is the set of all possible edges between $V$, that is, the arc set of the complete graph $G = (V,A)$.
Each component of a vector $\mathbf{x}$, namely $x_{ij}$, takes the value $1$ if there exists a transition from $i$ to $j$ in sequence $\mathbf{x}$, and $0$ otherwise. The CVRP routing problem with $m$ homogeneous vehicles, each with capacity $Q$, and a given demand $q_i$ for every stop $i \in V$, can then be represented by the following standard constraints [\cite{toth2014family}]:
\begin{align}
    & \sum_{j\in V\!,\: j\neq i} x_{ij} = 1 && i\in V \label{eqn:con_flow1}\\
    & \sum_{i\in V\!,\: i\neq j} x_{ij} = 1 && j\in V \label{eqn:con_flow2}\\
    & \sum_{j=1}^n x_{0j} \leq m \label{eqn:con_fleet}\\
    & \text{if} \ x_{ij}=1 \ \Rightarrow \ u_i + q_j = u_j && (i,j) \in A : j\neq 0,\, i\neq 0 \label{eqn:con_cap1}\\
    & q_i \leq u_i \leq Q && i\in V\backslash\{0\} \label{eqn:con_cap2}\\
    & x_{ij} \in \{0, 1\} && (i,j) \label{eqn:con_integ}\in A,
\end{align}
\noindent where constraints (\ref{eqn:con_flow1}) and (\ref{eqn:con_flow2}) impose that every stop other than the depot must be visited by exactly one vehicle and that exactly one vehicle must leave from each node. Constraint (\ref{eqn:con_fleet}) limits the number of routes to the size of the fleet, $m$. In constraint (\ref{eqn:con_cap1}), $u_j$ denotes the cumulative vehicle load at node $j$. The constraint plays a dual role---it prevents the formation of subtours, i.e., cycling routes that do not pass through the depot, and together with constraint (\ref{eqn:con_cap2}), it ensures that the vehicle capacity is not exceeded. While the model does not make explicit which stop belongs to which route, this information can be reconstructed from the active arcs $x_{ij}$ in the solution.

\subsubsection*{Objective function.}
For a first order Markov chain approximation, the joint probability over a daisy-chained sequence of stops $\mathbf{x} = \{0, s_1, s_2,s_3,\ldots s_n, 0\}$ decomposes into the following:
\begin{align}
\begin{split}
 \Prob(\mathbf{x})\ & \approx\ \Prob(X_0 = 0)\,\Prob(X_1 = s_1\, |\, X_0 = 0)\, \cdots\, \Prob(X_{n+1} = 0\,|\, X_{n} = s_n) \label{eq:probability} \\ 
 & =\ \Prob(X_0 = 0) \prod_{(i\rightarrow j)\, \in\, \mathbf{x}} \mathbf{Pr}\left(X_{t+1} = j \, |\, X_t = i \right).
 \end{split}
\end{align}
In our routing setting, by construction, the first stop is always the depot. Hence, we know that $\Prob(X_0 = 0) = 1$. In order to transform the above into an objective function over decision variables $x_{ij}$ we remark that $x_{ij} = 1 \Leftrightarrow (i\rightarrow j) \in \mathbf{x}$. We can now derive the following:
\begin{align}
\argmax_{\mathbf{x}}~ \Prob(\mathbf{x})\ & \approx\ \argmax_\mathbf{x}~ \Prob(X_0 = 0) \prod_{(i\rightarrow j)\, \in\, \mathbf{x}} \mathbf{Pr}\left(X_{t+1} = j \, |\, X_t = i \right) \nonumber \\
 & =\ \argmax_\mathbf{x}~ \prod_{(i\rightarrow j)\, \in\, \mathbf{x}} \mathbf{Pr}\left(X_{t+1} = j \, |\, X_t = i \right) \nonumber \\
 & =\ \argmax_x \prod_{(i,j)\, |\, x_{ij} = 1} \mathbf{Pr}\left(X_{t+1} = j \, |\, X_t = i \right)x_{ij} \nonumber \\
 & =\ \argmax_x \sum_{(i,j)\, \in\, A} log\,\mathbf{Pr}\left(X_{t+1} = j \, |\, X_t = i \right)x_{ij} \nonumber \\
 & =\ \argmin_x \sum_{(i,j)\, \in\, A} -log\,\mathbf{Pr}\left(X_{t+1} = j \, |\, X_t = i \right)x_{ij}.
\end{align}
If we define $\hat{p}_{ij} = -log\,\mathbf{Pr}\left(X_{t+1} = j \, |\, X_t = i \right)$, then we obtain the following \textbf{standard CVRP formulation}:
\begin{align}
    \min_{x} & \sum_{(i,j)\,\in\, A} \hat{p}_{ij} x_{ij} \label{eq:prob_first} \\
    \mbox{s.t. } & \mbox{ Constraints \eqref{eqn:con_flow1} - \eqref{eqn:con_integ}}. \nonumber
\end{align}
Hence, using $\hat{p}$ as the cost vector in the traditional VRP setting enables us to use any existing VRP solver to find the maximum likelihood routing of a given first order Markov model.


\vspace{1em} 
\noindent
We now consider the case of a \textbf{second order Markov chain}:
\begin{align}
\begin{split}
 \Prob(\mathbf{x})\, \approx & \ \Prob(X_0 = 0) \, \Prob(X_1 = s_1 \,|\, X_0 = 0) \, \Prob(X_2 = s_2 \,|\, X_0 = 0, X_1 = s_1)\,\cdots \\
                           &\ \qquad \Prob(X_{n+1} = 0 \,|\, X_{n-1} = s_{n-1}, X_{n} = s_n) \label{eq:probability2} \\ 
\end{split}
\nonumber \\
\begin{split}
&=\  \Prob(X_0 = 0)\,\Prob(X_1 = s_1\,|\,X_0 = 0) \prod_{(i\rightarrow j \rightarrow k) \,\in\, \mathbf{x}} \mathbf{Pr}\left(X_{t+1} = k \, |\, X_{t-1} = i, X_t = j \right).
 \end{split}
\end{align}
To construct a corresponding objective function over decision variables $x_{ij}$, we have to be more careful than in the first order case. More specifically, the second order Markov model includes the probabilities of transitions over the different vehicles, such as $(s_{1p} \rightarrow 0 \rightarrow s_{21})$. 
However, vehicles (and hence routes) are assumed to be homogeneous and therefore interchangeable. Hence, the ordering of the routes when constructing the daisy-chain is arbitrary. The last stop of one route should therefore not have any influence on the first stop of another route. We will hence ignore all transitions of the kind $(s_{1p} \rightarrow 0 \rightarrow s_{21})$ and instead use the first order transition probability from the depot: $(0 \rightarrow s_{21})$. These need to be estimated for the first transition $\Prob(X_1 = s_1 \,|\, X_0 = 0)$ anyway. This leads to the following derivation:
\begin{align}
\begin{split}
\argmax_\mathbf{x}~ & \Prob(X_0 = 0)\,\Prob(X_1 = s_1\,|\,X_0 = 0) \prod_{(i\rightarrow j \rightarrow k) \,\in\, \mathbf{x}} \mathbf{Pr}\left(X_{t+1} = k \, |\, X_{t-1} = i, X_t = j \right) \\ \nonumber
\approx\ & \argmax_\mathbf{x}~ \Prob(X_0 = 0) \prod_{(0\rightarrow i) \,\in\, \mathbf{x}} \Prob(X_{t+1} = i \,|\, X_t = 0) \prod_{\substack{(i\rightarrow j \rightarrow k) \,\in\, \mathbf{x}\\ j\neq0}} \mathbf{Pr}\left(X_{t+1} = k \, |\, X_{t-1} = i, X_t = j \right) 
\end{split}
\nonumber \\ 
\begin{split}
=\ & \argmin_x \sum_{(0,i)\, \in\, A} -log\,\Prob(X_{t+1} = i \,|\,X_t = 0)\,x_{0i} \\
         &\qquad \qquad \qquad + \sum_{\substack{(i,j) \,\in\, A,\, (j,k)\, \in\, A\\ j\neq 0}} -log\,\mathbf{Pr}\left(X_{t+1} = k \, |\, X_{t-1} = i, X_t = j \right)\,x_{ij}\,x_{jk}.
\end{split} 
\end{align}
%

\noindent
If we now define $\hat{p}_{ijk} = -log\,\mathbf{Pr}\left(X_{t+1} = k \, |\, X_{t-1} = i, X_t = j \right)$, and $\hat{p}_{ij} = -log\,\mathbf{Pr}\left(X_{t+1} = j \, |\, X_t = i \right)$ as before, then we obtain the following CVRP formulation:
\begin{align}
    \min_{x} &\sum_{(0,i)\, \in\, A} \hat{p}_{0i}x_{0i} + \sum_{\substack{(i,j)\in A \\ j\neq 0}}\sum_{\substack{(j,k)\in A \\ j\neq 0}} \hat{p}_{ijk} x_{ij} x_{jk} \label{eq:prob_second}\\
    \mbox{s.t.  } & \mbox{ Constraints \eqref{eqn:con_flow1} - \eqref{eqn:con_integ}}. \nonumber
\end{align}
Note how we are still using the same $x_{ij}$ decision variables. The objective function, however, is now a quadratic function and hence the problem becomes a Mixed Integer Quadratic Problem (MIQP). There are well-known techniques to linearize the quadratic term and obtain an MIP by introducing additional constraints and decision variables, e.g., by the McCormick inequalities [\cite{mccormick1976computability}].
Many mathematical programming solvers for MIQP also exist. However, it is reasonable to expect that optimizing with this objective function of the second order Markov chain will be computationally harder than optimizing with the linear objective function of the first order model.

We now explain how to learn the transition probability matrix from historical solutions (Section~\ref{probmatrixconst}), followed by different ways of weighing the instances  (Section~\ref{weighingschemes}). Finally, in Section~\ref{distancebasedprob} we discuss how to combine a learned probability matrix with a distance-based probability~matrix.

\subsection{Learning the transition probability matrix}
\label{probmatrixconst}

We now outline the main approach that we propose for estimating the transition probability matrix from historical solutions. We assume given a dataset 
${\cal H} = \{(V^t, m^t, z^t, {\mathbf x}^t)\}$
of instances with each instance a tuple where $t$ is a timestamp (e.g., a specific day in case of daily vehicle routing), $V^t$ is the set of stops served by the solution at timestamp $t$, $m^t$ is the number of vehicles used, ${\mathbf x}^t$ is the \textit{subjectively} optimal solution, that is, a solution created by an expert or the actual driven routes extracted from an on-board system, and $z^t$ are additional problem parameters such as the demand of every stop, or some other known constraint parameters.

Note that $V^t$, as well as $m^t$ and $z^t$, can change from instance to instance. The approach we propose assumes that while $V^t$ indeed changes from day to day, there will always be some overlap with other days, for example, because the set of stops is composed of \emph{regular} and occasional or \emph{ad hoc} stops. 



\subsubsection*{Probability estimation.}
The basic idea of our approach is to estimate all the conditional probabilities $\mathbf{Pr}(X_{n+1}\!=\! {j}\,|\, X_n\!=\!i)$ over the set of all stops in the data: $V^{all} = 
\bigcup_t V^t$.

The basic properties of conditional probability theory state that the conditional probability of moving from stop $i$ to stop $j$ is:
\[
\mathbf{Pr}(X_{n+1}\!=\! {j}\mid X_n\!=\!i) = \frac{\mathbf{Pr}(X_{n+1}\!=\!{j},\ X_n\!=\!i)}{\mathbf{Pr}(X_n \!=\! i)},
\]
with $\mathbf{Pr}(X_n\!=\!i) = \sum_k \mathbf{Pr}(X_{n+1}\!=\!{k},\ X_n\!=\!i)$.

To compute this, we will count how often two stops follow each other in the data. We denote by $\llbracket\, \cdot\, \rrbracket$ the Iverson bracket which returns the value $1$ if the statement inside the bracket evaluates to \textit{true}, and $0$ otherwise. We define the frequency of a transition $(i \rightarrow j)$ in dataset ${\cal H}$ as:
\begin{equation}
  f_{ij} = \sum_t
  \llbracket\, (i \rightarrow j) \in {\mathbf x}^t \,\rrbracket.
\end{equation}
Now we can empirically estimate the conditional probabilities from the data as follows:
 \begin{equation}
 \mathbf{Pr}(X_{n+1}\! =\! s_{j}\mid X_n\! =\! s_i)\ =\ \frac{f_{ij}}{\sum_k f_{ik}}.
 \end{equation}

\subsubsection*{Laplace smoothing.}
To account for the fact that the number of samples may be small, and some $f_{ij}$ may be zero, we can smooth the probabilities using the Laplace smoothing technique [\cite{johnson1932probability, chen1999empirical,ye2015method}]. Laplace smoothing reduces the impact of data sparseness arising in the process of building the transition matrix.
As a result of smoothing, these arcs are given a small, non-negative probability, thereby eliminating the zeros in the resulting transition matrix. 
Conceptually, with $\lambda$ as the smoothing parameter ($\lambda\!=\!0$ corresponds to no smoothing), we add $\lambda$ observations to each event. The conditional probability estimation therefore becomes: 
\begin{equation}
\mathbf{Pr}(X_{n+1}\! =\! s_{j}\mid X_n\! =\! s_i)\ =\ \frac{f_{ij} + \lambda}{\sum_k (f_{ik} + \lambda)}.
\end{equation}

\def\plus{\texttt{+}}
\begin{algorithm}[t]
  \caption{Estimating a first order transition matrix from historical instances \label{alg:estimate}}
  \textbf{Input:} A dataset ${\cal H} = \{(V^t, m^t, z^t, {\mathbf x}^t)\}$ where we will assume that the timestamps $t$ have been replaced by integer values $1\ldots|\,{\cal H}\,|$ in a way that respects the ordering of the timestamps; a weight $w_t$ per data instance, where the default value is $w_t=1$; and the Laplace smoothing parameter $\lambda\geq0$. 
  \begin{enumerate}[leftmargin=*,topsep=0pt]
    \item Determine the total set of stops 
    $V^{all} = \bigcup_t V^t$
    and let $\mu = |\, V^{all}\,|$.
    \item For each $(V^t, m^t, z^t, {\mathbf x}^t) \in {\cal H}$, construct an adjacency matrix $\mathbf{A}^t_{\mu\,\times\, \mu} = [a^t_{ij}]$, where
    \begin{equation}
      a^t_{ij}= \llbracket\, (s_i,s_j)\in {\mathbf x}^t \,\rrbracket.
    \end{equation}
    \item Build the arc transition frequency matrix $\mathbf{F}_{\mu\,\times\, \mu}$ with the weights $w_t$ and the adjacency matrices constructed in Step 2:
    \begin{equation}
    \label{eqn:algo_weights}
    \mathbf{F} = \sum_t w_t\mathbf{A}^t.
    \end{equation}
    \item Apply the Laplace smoothing technique to $\mathbf{F}_{\mu\,\times\, \mu} = [f_{ij}]$ to get the final probability estimates $\mathbf{\hat{P}}_{\mu\,\times\, \mu} = [\hat{p}_{ij}]$:
    \begin{equation}
    \label{eqn:transmat_elem}
    \hat{p}_{ij} = \frac{f_{ij}+\lambda}{\sum_{k}{(f_{ik}+\lambda)}}. 
    \end{equation}
  \end{enumerate}
  \textbf{Output:} Transition matrix $\mathbf{\hat{P}}_{\mu\times\, \mu} = [\hat{p}_{ij}]$, where $\hat{p}_{ij}\ =\ \mathbf{Pr}(X_{n+1} = s_{j}\,|\, X_n = s_i)$.
\end{algorithm}

\subsubsection*{First-order estimation algorithm.}
\textbf{Algorithm \ref{alg:estimate}} shows the algorithm for estimating the first order probability transition matrix. The dimension of the matrix, that is, the size of the total set of unique stops, is determined in Step 1. In Step 2, an adjacency matrix is constructed for each historical instance. A frequency matrix is constructed in Step 3 by computing the (weighted) sum of all the adjacency matrices (Eq.~\eqref{eqn:algo_weights}). By default, $w_t = 1$ for all instances; more advanced schemes will be discussed in Section \ref{weighingschemes}. Finally, during normalisation in Step 4, Laplace smoothing is applied if $\lambda > 0$.

\subsubsection*{Second-order estimation algorithm.}
The second order transition matrix can be analogously constructed by building a three-dimensional matrix $\mathbf{A}^t_{\mu\,\times\, \mu\,\times\, \mu} = [\,a^t_{ijk}\,]$ in Step 2, where $a^t_{ijk}= \llbracket\, (s_i,s_j,s_k) \in {\mathbf x}^t \,\rrbracket$. Laplace smoothing is then applied by dividing each element of the frequency matrix $\mathbf{F}_{\mu\,\times\, \mu\,\times\, \mu} = [\,f_{ijk} :=  \sum_t
  \,\llbracket\, (i \rightarrow j), \, (j \rightarrow k) \in {\mathbf x}^t \,\rrbracket\,]$ with the row sum to obtain the transition matrix $\mathbf{\hat{P}}_{\mu\,\times\, \mu\,\times\, \mu} = [\,\hat{p}_{ijk}\,],$ where
\begin{align*}
    \hat{p}_{ijk}\ &=\ \mathbf{Pr}(X_{n+1} = s_{k}\mid X_{n-1}=s_i, X_n = s_j)
          =\ \frac{f_{ijk}+\lambda}{\sum_{l}{(f_{ijl}+\lambda)}}.
\end{align*}

\noindent
We also estimate the conditional probabilities of leaving from the depot:
\begin{align*}
    \hat{p}_{0i} = \frac{f_{0j}+\lambda}{\sum_{k}{(f_{0k}+\lambda)}}.
\end{align*}
With these estimates, we can use the above-defined optimisation formulation to find the maximum likelihood VRP solution.

\subsection{Every day is different}
\label{weighingschemes}
The estimation method described above assumes that each instance is equally important, and that the VRP solution of each instance is independently drawn from the same distribution. However, we know that this is not entirely true: human planners learn from day to day, with more recent experiences typically closer to mind. Furthermore, because the set of stops changes each day, the similarity of the current instance to previous instances may also change how choices will be made.

We identify two types of \textit{context} that change the importance of previous instances---the temporal context and the similarity context. The temporal context is known in machine learning as \textit{concept drift}~[\cite{gama2014survey}]. The concept of similarity is central in many machine learning and data mining approaches. We now discuss two modifications of the above transition probability estimation algorithm that will take these contexts into account.

To this end, we assume that we are currently at timestamp $T.$ Also, we are given a set of stops $V^T$, number of vehicles $m^T$, and other parameters $z^T.$ That is, we have a new unseen tuple $(V^T, m^T, z^T)$ for which we are to determine the corresponding ${\mathbf x}^T$. We also have a set of historical instances until timestamp $T$ that we denote with ${\cal H} = \{(V^t, m^t, z^t, {\mathbf x}^t)\}_{t=1}^{T-1}.$

 

To take the temporal and similarity aspects into account, we will define a \textit{prior} on each historical training instance, based on the current tuple $(V^T, m^T, z^T)$. More specifically, we use a \textit{weighing} scheme where we define a weight $w_t$ for each historical instance in ${\cal H}$ based on the properties of the current tuple.

Table \ref{tab:schemes} provides an overview of the three types of prior: uniform, time-based, and similarity-based, with other possible variations within each type.

\begin{table}[t]
\begin{scriptsize}
\begin{center}
\caption{An overview of the proposed weighing schemes}
\label{tab:schemes}
{\tabcolsep=0pt\def\arraystretch{1.5}
\begin{tabular}{@{}llll@{}}
\toprule
    Name & \qquad Weights & \quad\: Squared Weights & \quad\: Exponential Weights (EXP)\\ \cmidrule(){1-1}\cmidrule(lr){2-2}\cmidrule(){3-3}\cmidrule(l){4-4}
    Uniform (UNIF) & \qquad $w_t=1$ & \quad\: --- \\
    Time-based (TIME) & \qquad $w_t = t/T$ & \quad\: $w_t = (t/T)^2$ & \quad\: $w_t = \alpha(1-\alpha)^{T-t}$\\
    Similarity-based (SIMI) \quad & \qquad $w_t = J(V^t,V^T)$ \quad \quad & \quad\: $w_t = J(V^t,V^T)^2$ \\ \bottomrule
\end{tabular}}
\end{center}
\end{scriptsize}
\end{table}


\subsubsection*{Time-based weighing.}
In machine learning, it is well known that temporally ordered data can have \textit{concept drift}~[\cite{gama2014survey}], that is, the underlying distribution can change over time.
To account for this, we can use a time-based weighing scheme where older instances are given smaller weights, and newer instances larger ones. Assuming the timestamps are (replaced by) integer values in a way that respects the same ordering, we can weigh the instances as:
\begin{align}
w_t = \frac{t}{T}. 
\end{align}
This assumes a \emph{linearly} increasing importance of instances, with the oldest (first) instance having weight $1/T$ and the latest instance weight $(T-1)/T$.
We can increase the importance of the newer instances by squaring the weights: $w_t = (t/T)^2$ or more generally, $w_t = (t/T)^a$ for some value $a$.

\subsubsection*{Exponential smoothing.}
A further amplification of the importance of more recent instances can be done by considering an exponential weighing scheme, a popular approach in time series analysis and forecasting [\cite{cox1961prediction,harrison1967exponential}].

In principle, exponential smoothing uses a weighted average of the most recent observations and the forecasted data.
In general, let $\hat{f}_{T}$ be the estimated data up to, but not including time period $T$. Then, using the data $f_T$ of the current timestamp, the following gives the forecast for the next time period:
\begin{align}
\hat{f}_{T+1} = \alpha f_T + (1-\alpha) \hat{f}_{T},
\label{eq:exp_smoothing}
\end{align}
where the smoothing parameter $\alpha \in(0,1)$ is a weight assigned to the most recent data.
Furthermore, the expansion of Eq.~\eqref{eq:exp_smoothing} yields
\begin{align}
\hat{f}_{T+1} = \sum_{t=1}^T \alpha (1- \alpha)^{T-t} \ f_{T-t}.
\label{eq:exp_smoothing_expansion}
\end{align}
This weighing scheme is called exponential smoothing because the weights in Eq.~\eqref{eq:exp_smoothing_expansion} decline exponentially with time.

We can apply the same exponential smoothing on the frequency matrices, resulting in $\mathbf{F} = \sum_t \alpha (1- \alpha)^{T-t} \ \mathbf{A}^t$ and hence:
\begin{align}
w_t = \alpha(1-\alpha)^{T-t}. 
\end{align}

\subsubsection*{Similarity-based weighing.}
The stops in each instance typically vary, and the presence or absence of different stops can lead to different decision behaviors. To account for this, we can use the similarity between the set of stops of the current instance and the set of stops of each historical instance as prior. 
The goal is to assign larger weights to training instances that are more similar to the test instance, and smaller weights if they are less similar.

The similarity of two stop sets can be measured using the Jaccard similarity coefficient. The Jaccard similarity of two sets is defined as the size of the intersection divided by the size of the union of the two sets:
\begin{align}
    J(A,B) = \frac{|A\cap B|}{|A\cup B|}
\end{align}
for two non-empty sets $A$ and $B$. The Jaccard similarity coefficient is always a value between $0$ (no overlapping elements) and $1$ (exactly the same elements). 

Hence, we can use the following distance-based Jaccard similarity weighing:
\begin{align}
w_t = J(V^t,V^T). 
\end{align}
To further amplify the differences in the weights, we can also use the squared Jaccard coefficient $w_t = J(V^t,V^T)^2$ or in general, $w_t = J(V^t,V^T)^a$.

\subsection{Mixing distances and preferences}
\label{distancebasedprob}

%

We have discussed learning preferences from historical solutions and how to optimize over them, leading to the most likely VRP solution in expectation. However, this uses \textit{only} the historical preferences but never reasons at the level of distances traveled (kilometers). Using only the probability matrix can hence lead to purely \textit{mimicking} the human behaviour, instead of reasoning and optimizing over both preferences and the impact on the driven kilometers.



To face this issue, we define a \emph{distance-based probability} distribution based on the \emph{softmax} function on the distances between stops. The larger the distance between stops $i$ and $j$, the shorter is the probability of that transition. This probability is defined as:

\begin{align}
\begin{split}
\hat{d}_{ij}
 &\ =\ \mathbf{Pr}_{dist}(X_{n+1}\! =\! s_{j}\mid X_n\! =\! s_i)
\ =\  \,\frac{e^{-d_{ij}}}{\sum_k e^{-d_{ik}}}.
 \label{eqn:invcost}
\end{split}
\end{align}
The main goal is to have a comparable measure between the driver's preferences and the cost of driving long distances. In Appendix~\ref{append:distance}, we show that this definition of distance-based probabilities would produce, under some mild conditions, the same set of solutions as the standard CVRP formulation with the objective of minimizing the total distance.



\subsubsection*{Combining transition probability matrices.}
Given transition probability matrices $\mathbf{\hat{P}}=[\hat{p}_{ij}]$ and $\mathbf{\hat{D}}=[\hat{d}_{ij}]$, we can take the convex combination as follows:
\begin{equation}
    \hat{c}_{ij} = \beta \hat{p}_{ij} + (1-\beta) \hat{d}_{ij}.
    \label{eq:combinig_dis}
\end{equation}
Taking $\beta\!=\!1$ corresponds to using purely the history-based transition probabilities, while $\beta\!=\!0$ will use only distance-based probabilities, with values in between resulting in a combination of the two probabilities.

Note that this approach places no conditions on how the history-based transition probability matrix $\mathbf{\hat{P}}=[\hat{p}_{ij}]$ is computed, and hence is compatible with Laplace smoothing and weighing during the construction of $\mathbf{\hat{P}}$.

\section{Experiments} \label{s:expts}

\subsubsection*{Description of the Data.}
The historical data used in the experiments consist of daily routings collected within a span of nine months. The routings were generated by the route planners and used by the company in their actual operations. Each data is a numbered instance and the entire data is ordered by time. 

Data instances are grouped by day-of-week including Saturday and Sunday. This grouping mimics the operational characteristic of the company. The entire data set is composed of 201 instances, equivalent to an average of 29 instances per weekday. Capacity demand estimates for each stop were provided by the company.
\subsubsection*{Data Visualization.} 
\begin{figure}[htpb]
    \centering
    \begin{minipage}[b]{0.45\textwidth}
          \includegraphics[width=\linewidth]{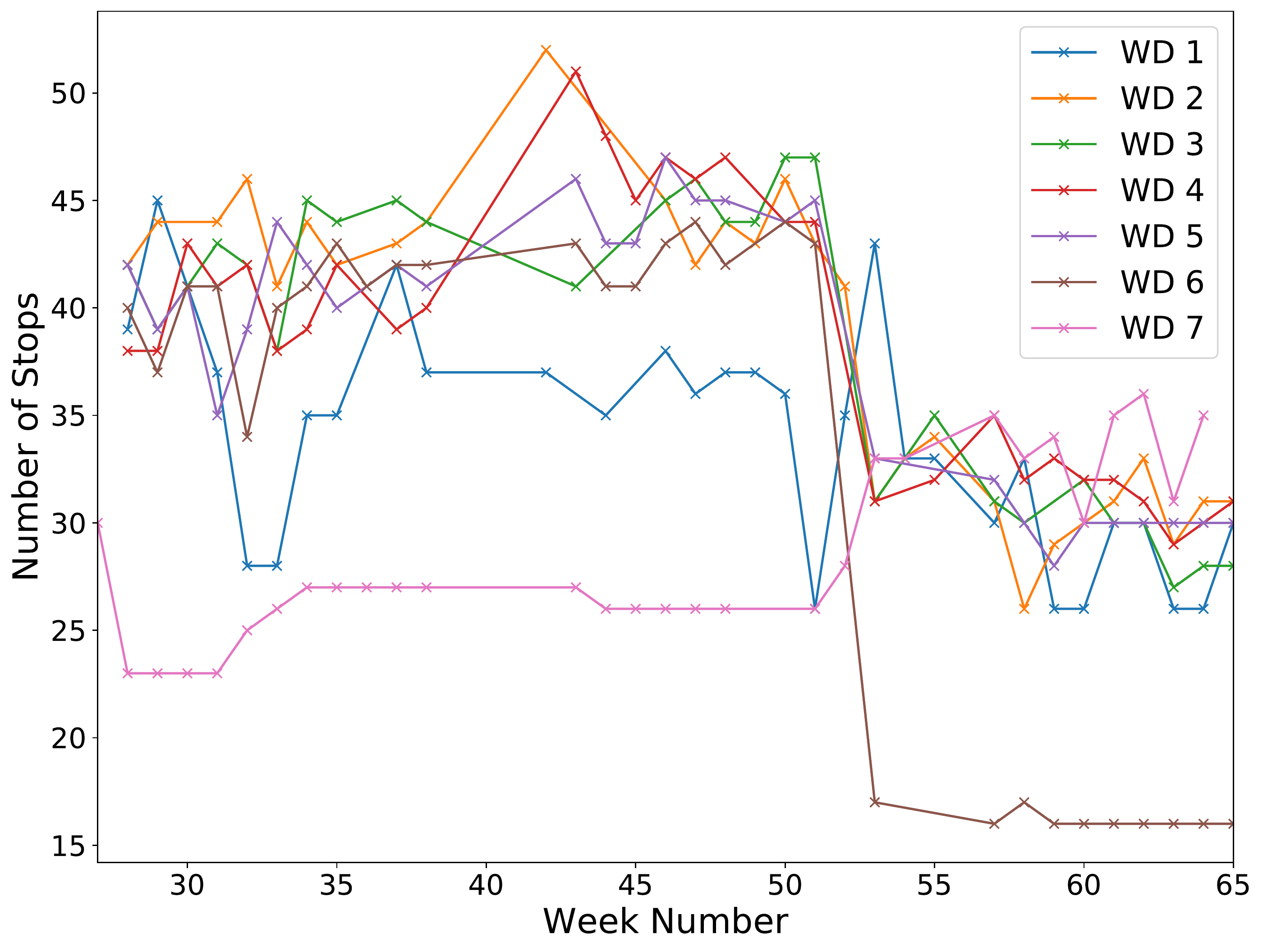}
          \caption{No. of stops by weekday (WD)}
          \label{fig:visualization}
    \end{minipage}
    \begin{minipage}{0.02\textwidth}
        $ $
    \end{minipage}%
    \begin{minipage}[b]{0.45\textwidth}
          \includegraphics[width=\linewidth]{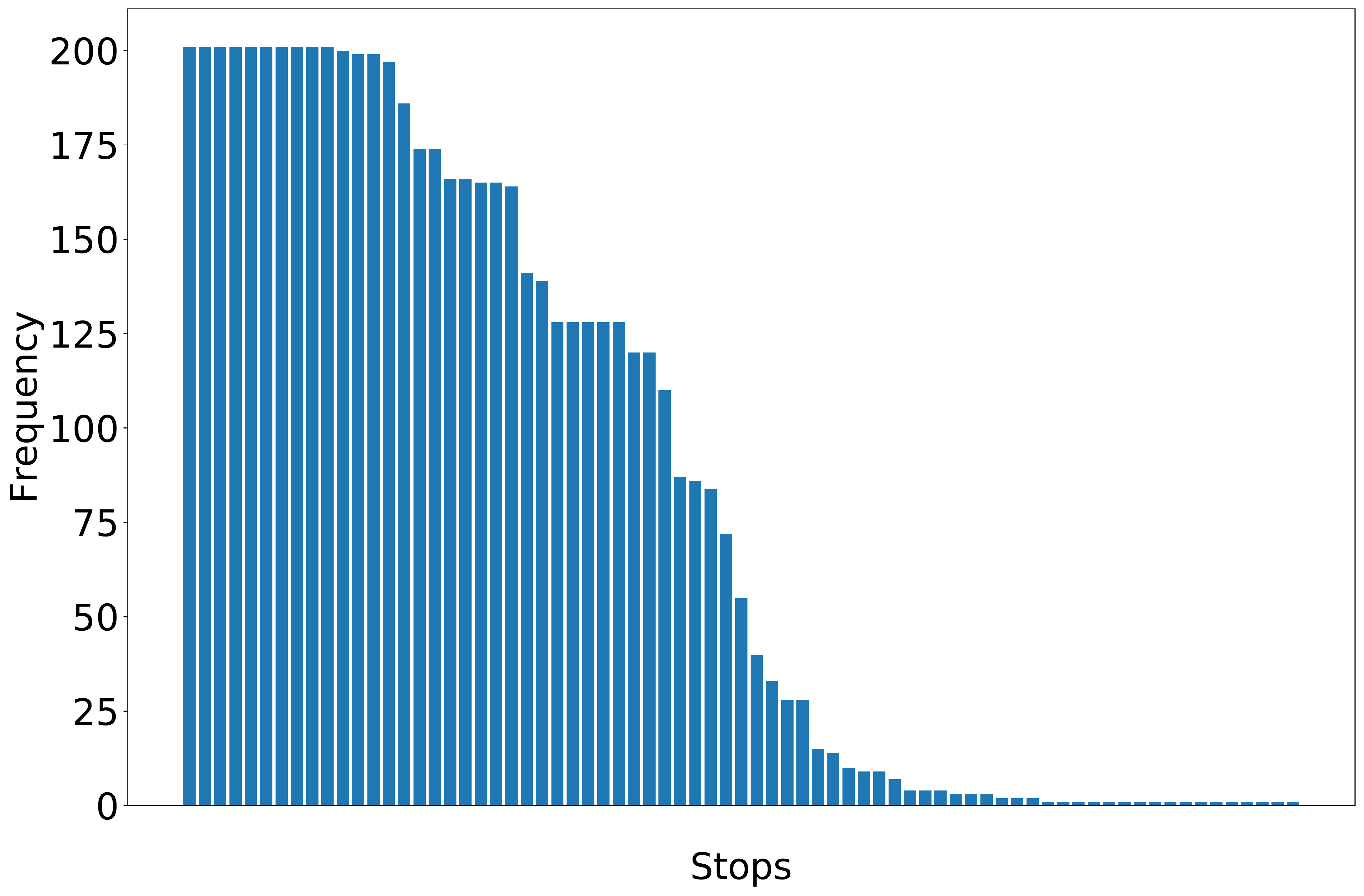}
          \caption{Frequency of Stops Across Data}
          \label{fig:stopfrequency}
    \end{minipage}
\end{figure}

\textbf{Figure \ref{fig:visualization}} shows the number of stops served per weekday during the entire experimental time period. A \emph{concept drift} is clearly discernible starting Week 53, where a decrease in stop set size occurs. 
An average of 9 vehicles servicing 35 stops are used per instance in the data before drift, and 6 vehicles (25 stops) for the 73 instances after drift. This observation has prompted us to make explicit whenever we are using all the data from the \emph{entire} period, or only data from the period \emph{before the drift.}

The number of times each of the customer stops has appeared in the historical data is shown in \textbf{Figure \ref{fig:stopfrequency}}, which exhibits a mix of regular and ad hoc (occasional) stops. At the extremes, 14 out of the 73 unique stops (19.2\%) have been serviced more than 195 (out of 201) times, while 30 (41.1\%) have appeared in only ten or less instances.

\subsubsection*{Evaluation Methodology.}
\label{datausage}
In a traditional machine learning setup, the dataset is split into a training set and a test set. The training set is used for training, and the test set for evaluation. This is a batch evaluation as all test instances are evaluated in one batch. The best resulting model 
is then deployed. 

Our data, however, has a temporal aspect to it, namely the routing is performed everyday. Hence, each day one additional training instance becomes available, allowing us to \textit{incrementally} grow the data and learn from it. Indeed, human planners also learn from prior experience and expand their knowledge day by day.

As this is the most sensible way in which such a system would be used, the system also has to be evaluated in this way, i.e., by \textbf{incremental evaluation}. The incremental evaluation procedure is depicted in \textbf{Algorithm 2} and the breakdown of the entire data set after the initial train-test split is shown in \textbf{Table \ref{tab:split}}.

        \begin{table}[h!]
        \centering
        \begin{scriptsize}
        \vfill
        \caption{Initial training and test set sizes after $75\%-25\%$ split} 
        {
        \tabcolsep=0pt\def\arraystretch{1}
        \begin{tabularx}{0.6\textwidth}{XXXXX}
       \toprule
            & \multicolumn{2}{c}{Before drift}   
            & \multicolumn{2}{c}{Entire data} 
            \tabularnewline \cmidrule(lr){2-3}\cmidrule(l){4-5}
            WD & Train & Test & Train & Test \tabularnewline \midrule
            1 & 14 & 5 & 23 & 7 \tabularnewline
            2 & 12 & 5 & 21 & 7 \\
            3 & 11 & 5 & 19 & 7 \\
            4 & 13 & 5 & 22 & 7 \\
            5 & 14 & 5 & 22 & 7 \\
            6 & 14 & 5 & 22 & 7 \\
            7 & 15 & 5 & 23 & 7 \\ \bottomrule
            Total & 93 & 35 & 152 & 49 \\ \bottomrule
        \end{tabularx}
        }
        \vfill
        \label{tab:split}
        \end{scriptsize}
        \end{table}

\begin{algorithm}[htpb]
  \caption{Training and testing with an incrementally increasing training set}
  \textbf{Input:} A dataset ${\cal H} = \{(V^t, m^t, z^t, {\mathbf x}^t)\}$ with timestamps $t$  as in Algorithm~\ref{alg:estimate}.
  \begin{enumerate}[leftmargin=*,topsep=0pt]
    \item Start from an initial $\eta$ training instances, e.g., $\eta=\lfloor\, 0.75\,|{\cal H}| \,\rfloor$ for a $75\%-25\%$ initial split. 
    \item For $\sigma=\eta+1,\ldots,|{\cal H}|\!-\!1$ do:
    \begin{itemize}[label={}, topsep=0pt]
    \item[2.1.] Build the probability transition matrix $\mathbf{\hat{P}}^\sigma$ on ${\cal H^\sigma} =  \{(V^t, m^t, z^t, {\mathbf x}^t)\}_{t <\sigma} $ using Algorithm \ref{alg:estimate}.
    \item[2.2.] Add to $\mathbf{\hat{P}}^\sigma$ all stops in $V^\sigma$ that are not in ${\cal H^\sigma}$, with uniform probability.
    \item[2.3.] Solve CVRP (\ref{eq:prob_first}) using $\mathbf{\hat{P}}^\sigma$ as in Section~\ref{sec:mlrouting}.
    \item[2.4.] Evaluate the CVRP solution against ${\mathbf x}^\sigma$.
    \end{itemize}
  \end{enumerate}
\end{algorithm}

When making a comparison of the prediction accuracy of the proposed schemes, we evaluate the performance using two evaluation measures based on two properties of a VRP solution, namely the grouping of stops into routes ({\it Route Difference}) and the resulting chosen arcs ({\it Arc Difference}).


\emph{Route Difference} (RD) indicates the percentage of stops that were incorrectly assigned to a different route. Intuitively, RD may be interpreted as an estimate of how many moves between routes are necessary when modifying the predicted solution to match the actual grouping of stops into routes. To compute route difference, a pairwise comparison of the routes contained in the predicted and test solution is made. The pair of routes with the smallest difference in stops is greedily selected without replacement. Incorrectly assigned stops are counted and the total number is taken as RD. The percentage is taken by dividing RD by the total number of stops in the whole routing.

\emph{Arc}~\emph{Difference} (AD) measures the percentage of  arcs traveled in the actual solution but not in the predicted solution. AD is calculated by taking the set difference of the arc sets of the test and predicted solutions, then dividing the value by the total number of arcs in the routing. Correspondingly, AD gives an estimate of the total number of modifications needed to \emph{correct} the predicted solution.
\subsubsection*{Experimental Setup.}
The numerical experiments were performed using Python 3.7.4 and the CPLEX 12.9 solver with the default setting, on a Lenovo ThinkPad X1 Carbon with an Intel Core i7 processor running at 1.8GHz with 16GB RAM. 
 The Laplace smoothing parameter ($\lambda$), convex combination parameter ($\beta$) and exponential smoothing parameter ($\alpha$) take default values  $\lambda=1$, $\beta=1$, and $\alpha=0.7$.
Following the weighing scheme overview in \textbf{Table \ref{tab:schemes}}, we denote uniform weighing by UNIF, TIME for time-based, EXP for exponential time-based weighing and SIMI for similarity-based weights. TIME2 and SIMI2 indicate squared weights. Furthermore, whenever necessary, we use superscripts ($^1$) and ($^2$) to distinguish when a weighing scheme is applied using the first or the second order model, respectively. For instance, UNIF$^1$ indicates uniform weighing using the first order model.  


\begin{figure}[htpb]
\centering
\begin{subfigure}{.45\textwidth}
  \centering
        \includegraphics[width=0.99\linewidth]{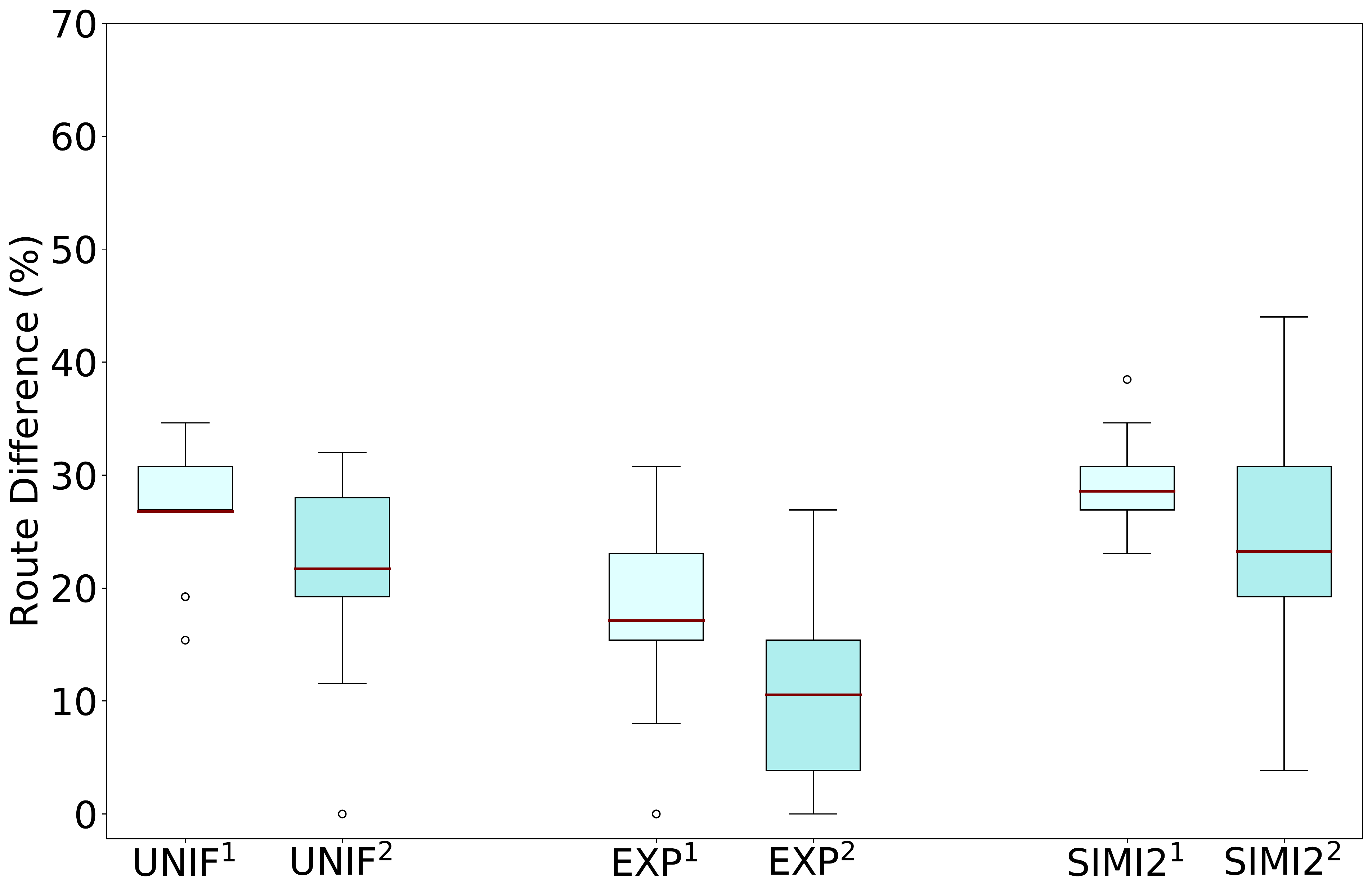}
\end{subfigure}
    \begin{minipage}{0.08\textwidth}
        \hspace{0.1cm}
    \end{minipage}%
\begin{subfigure}{.45\textwidth}
\centering
        \includegraphics[width=\linewidth]{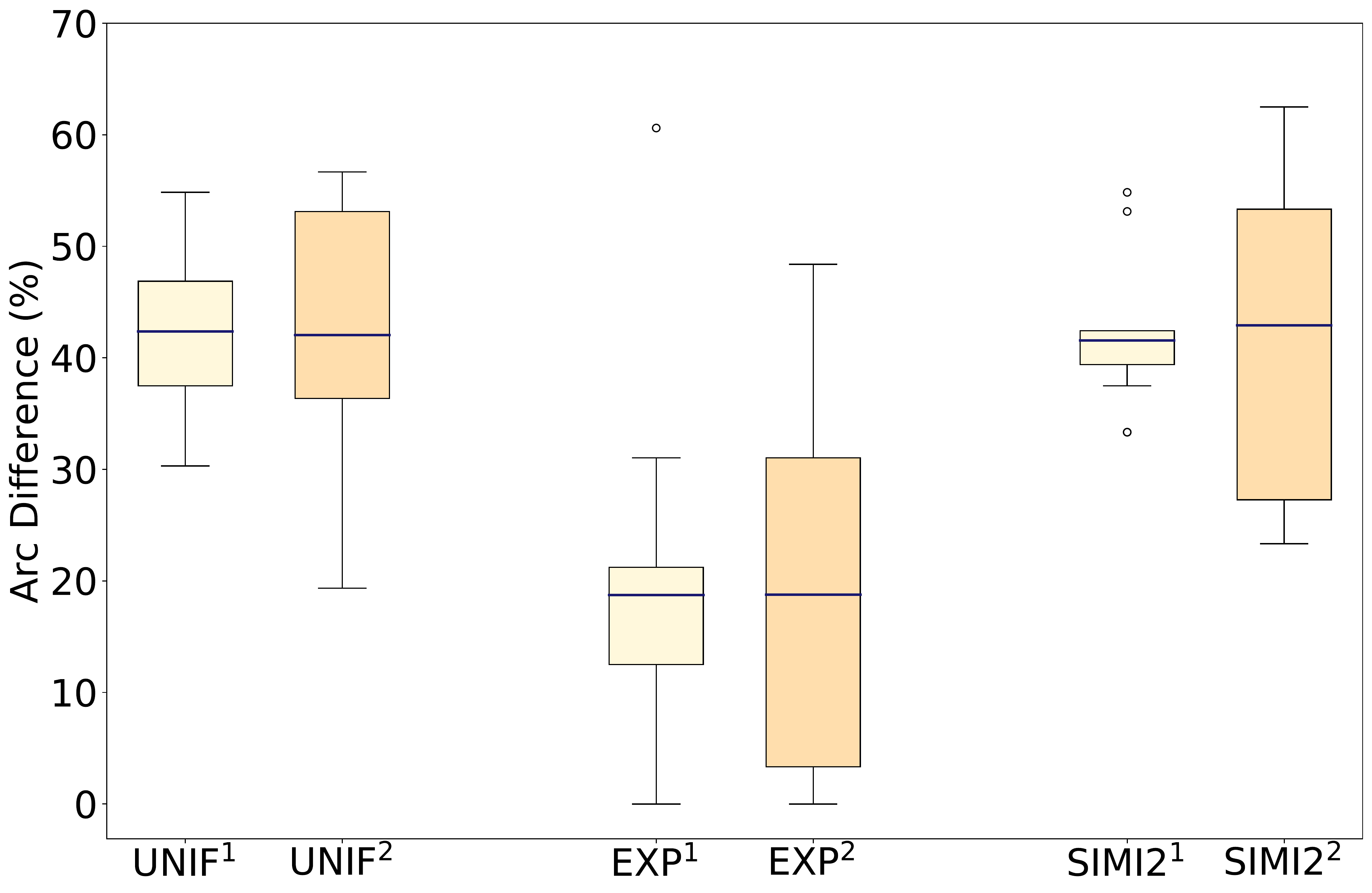}
\end{subfigure}

$ $

$ $

\begin{scriptsize}
{\tabcolsep=3pt\def\arraystretch{1.5}
\begin{tabular}{l|ll|ll|ll}
Weighing Scheme & \multicolumn{1}{c}{UNIF$^1$} & \multicolumn{1}{c|}{UNIF$^2$} & \multicolumn{1}{c}{EXP$^1$} & \multicolumn{1}{c|}{EXP$^2$} & \multicolumn{1}{c}{SIMI2$^1$} & \multicolumn{1}{c}{SIMI2$^2$} \\ \hline
Avg Computation Time (s) & 0.1016 & 4355.75 & 0.1455 & 2055.33 & 0.0922 & 9101.41 
\end{tabular}}

$ $

\caption{First order ($^1$) versus second order ($^2$) model (data from entire period)}
\label{fig:onevstwo}
\end{scriptsize}

\end{figure}

\subsection{First order versus second order model} We tested the performance of our models on a subset, composed of three weekdays, of all data from the entire period using three schemes (UNIF, EXP, SIMI2)---one from each type of prior: uniform, time-based and similarity-based. From the results shown in (\textbf{Fig. \ref{fig:onevstwo}}), we see that the second order model consistently produced better results in terms of route difference. However, this is not the case for arc difference, where aside from there being almost no change in prediction accuracy
, the variance also increased. We assert that more data points are necessary in order to better evaluate the second order model. 

Looking at the computation times, we see a considerable disparity between the two models. While schemes using the first order formulation resulted to subsecond computation times, using the fastest scheme in the second order model, EXP$^2$, took more than 2000 seconds for \emph{each} test instance to be solved to optimality.  We can hence remark that the slight improvement in route difference is, for practical purposes, reduced by the substantially longer running time of the second order model. 

For the succeeding experiments, we therefore focus our attention on further investigating the effects of the first order formulation.

\subsection{Evaluation of weighing schemes in the first order model}
\begin{figure}[t]
    \centering
    \begin{minipage}{0.4\textwidth}
        \centering
        \includegraphics[width=\linewidth]{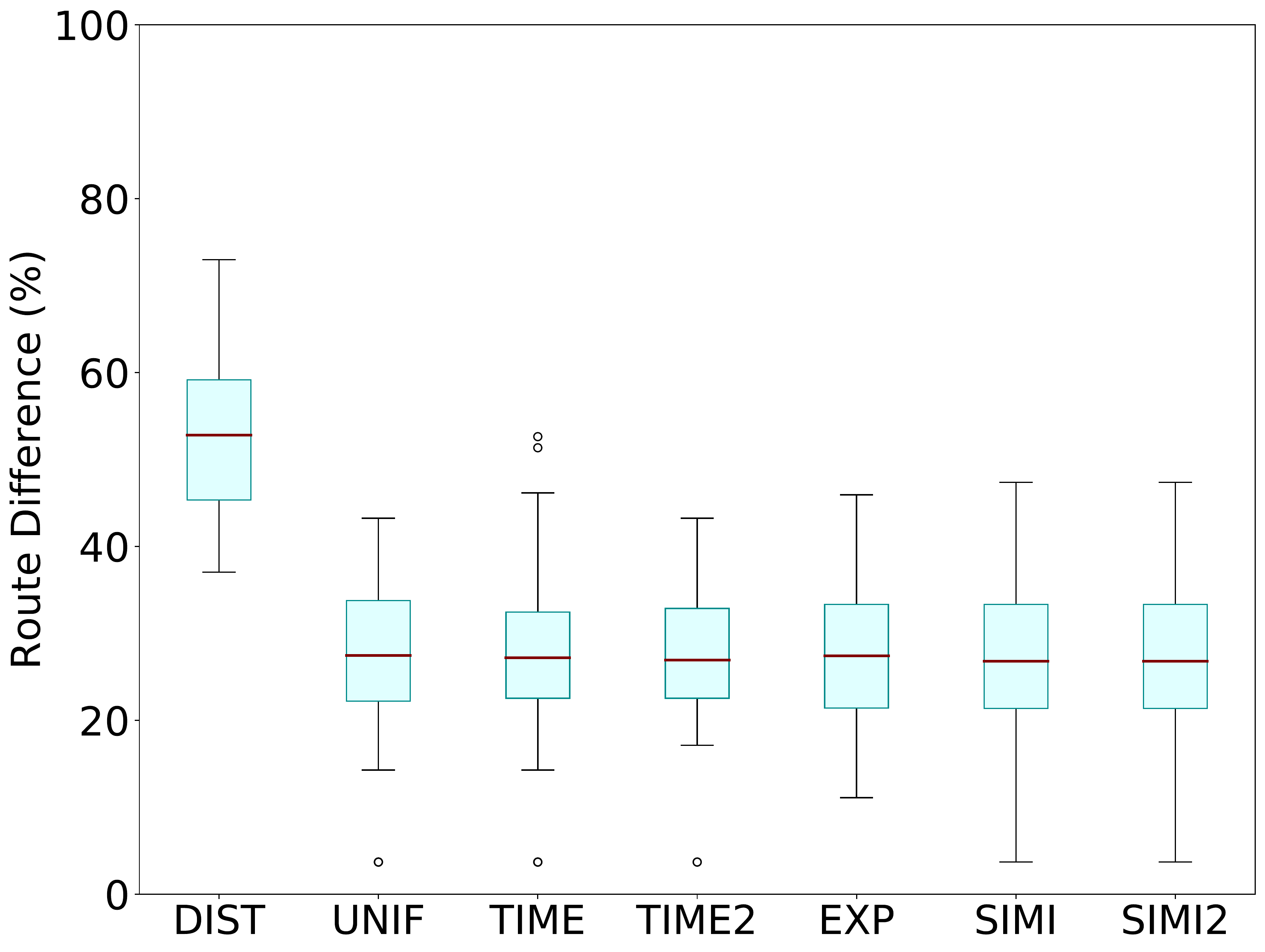}
    \end{minipage}%
    \begin{minipage}{0.08\textwidth}
        \hspace{0.2cm}
    \end{minipage}%
    \begin{minipage}{0.4\textwidth}
        \centering
        \includegraphics[width=\linewidth]{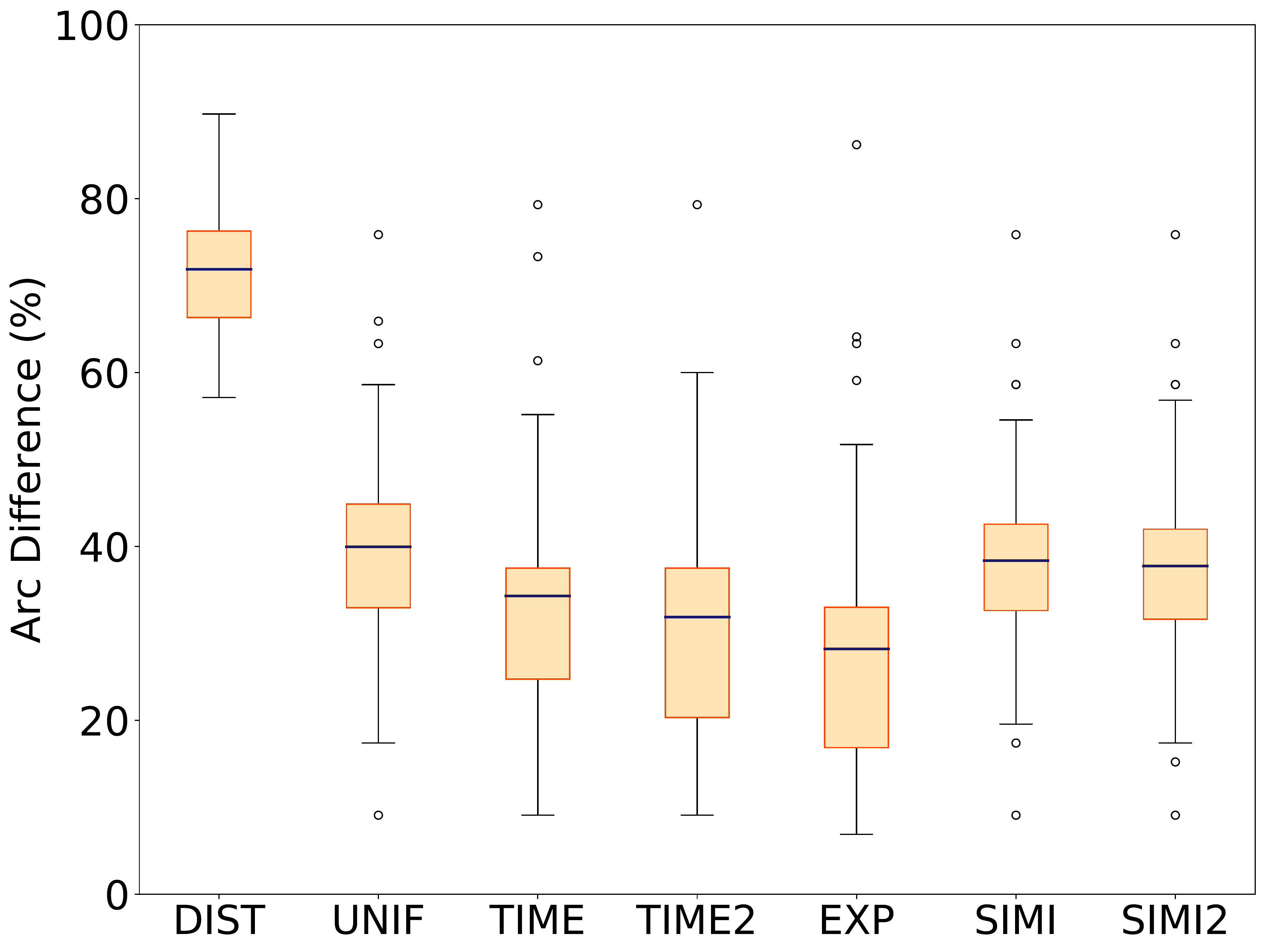}
    \end{minipage}
    \caption{Route and arc difference (period before drift)}
    \label{fig:BeforeDrift}
$ $

    \begin{minipage}{0.4\textwidth}
        \centering
        \includegraphics[width=\linewidth]{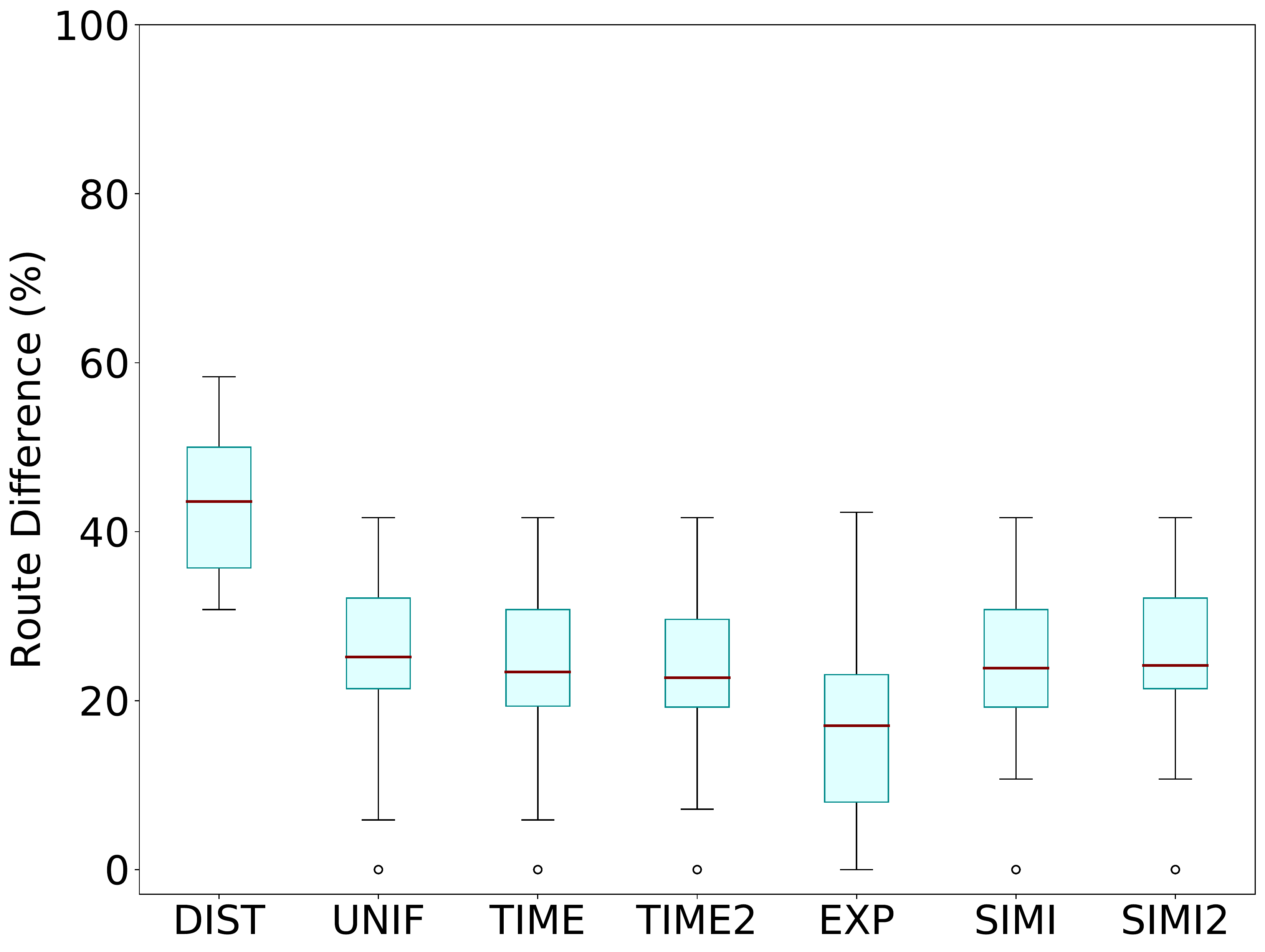}
    \end{minipage}%
    \begin{minipage}{0.08\textwidth}
        \hspace{0.2cm}
    \end{minipage}%
    \begin{minipage}{0.4\textwidth}
        \centering
        \includegraphics[width=\linewidth]{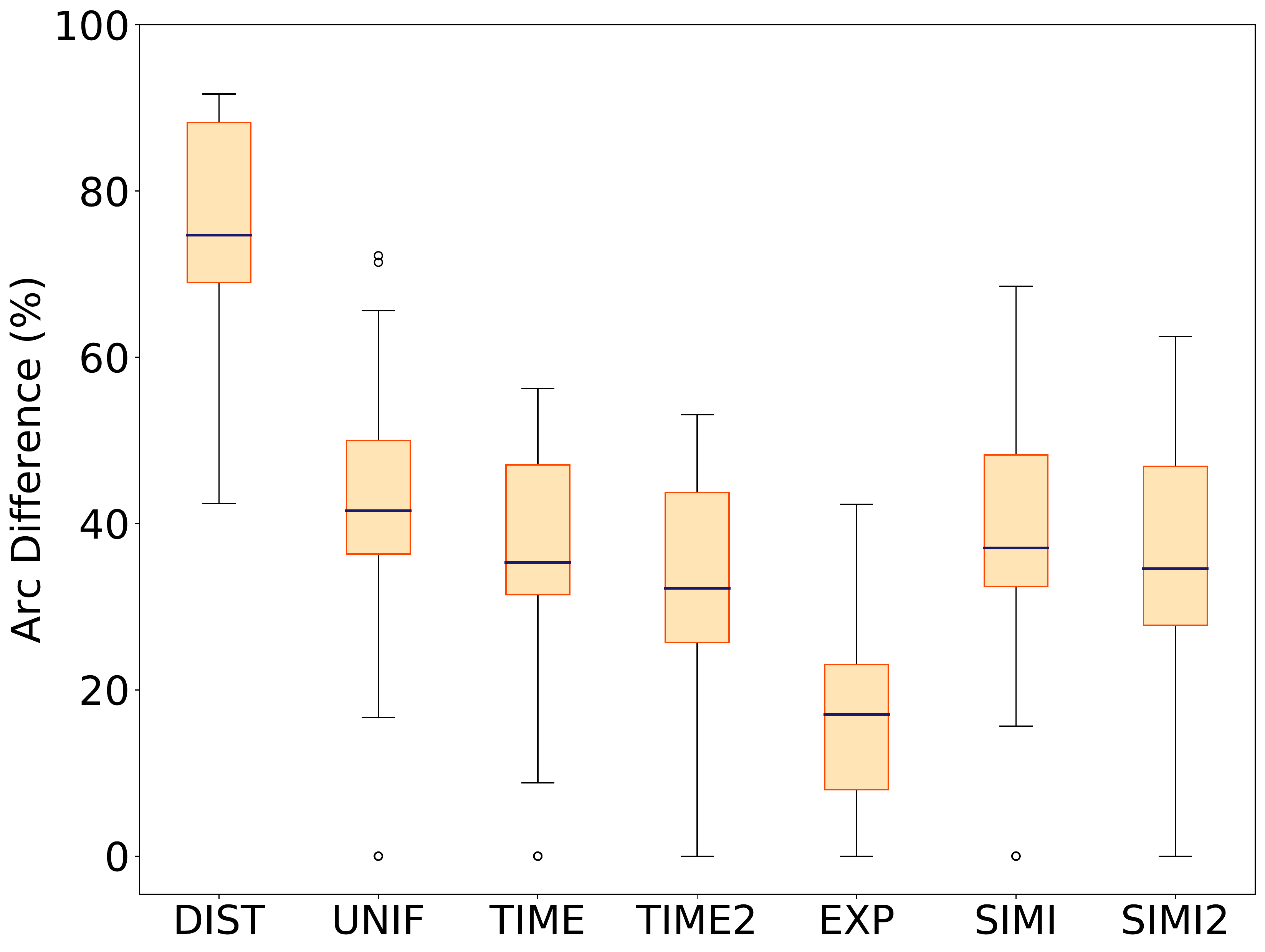}
    \end{minipage}
    \caption{Route and arc difference (entire period)}
    \label{fig:AllData}
\end{figure}

In the next two experiments (\textbf{Fig. \ref{fig:BeforeDrift}} - \textbf{\ref{fig:AllData}}), we do a wider comparison by investigating the performance of all our proposed weighing schemes (see \textbf{Table \ref{tab:schemes}})---UNIF, TIME, SIMI, and EXP, and TIME2 and SIMI2---in the first order model. 


\textbf{Fig. \ref{fig:BeforeDrift}} is on data before the drift (up to week 53 in \textbf{Fig. \ref{fig:visualization}}). It shows that all the proposed schemes produced better estimates than DIST. 
There appears to be no significant difference in the results in terms of route difference. For arc difference, however, using time-based weighing (TIME, TIME2, EXP) considerably improved the solutions given by UNIF. Among all schemes, EXP gave the most accurate predictions. Hence, it can be deduced that more recent routings are more relevant when making choices in this case.

Results on data from the entire period (\textbf{Fig. \ref{fig:AllData}}) exhibit a slightly different behavior. As before, all the schemes outperformed DIST. 
Notably in terms of arc difference, both the time-based and similarity-based schemes significantly outperformed UNIF. Also, in most cases, the schemes with the squared weights (TIME2, SIMI2) performed better than their counterparts (TIME, SIMI). As before, EXP gave the most accurate predictions among all schemes.  

\subsection{How the weighing schemes handle concept drift}
\label{drift_expt}
The two succeeding experiments were conducted to observe the performances of the proposed schemes during a concept drift. Our motivation is that we want to determine how the prediction quality of each scheme evolves when there is a sudden change in data structure.


\begin{figure}[t]
    \centering
    \begin{minipage}{0.4\textwidth}
        \centering
        \includegraphics[width=\linewidth]{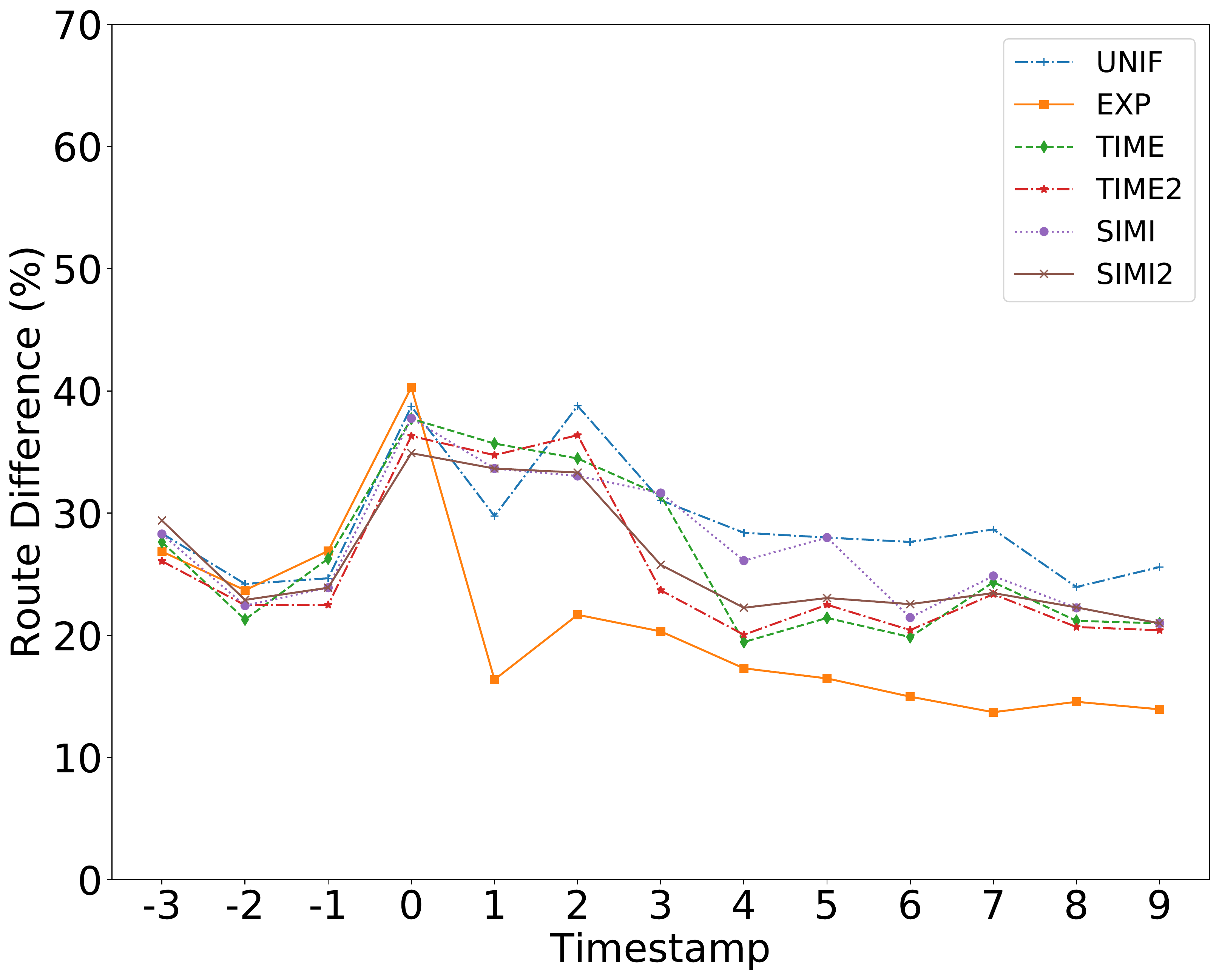}
    \end{minipage}%
    \begin{minipage}{0.08\textwidth}
        \hspace{0.1cm}
    \end{minipage}%
    \begin{minipage}{0.4\textwidth}
        \centering
        \includegraphics[width=\linewidth]{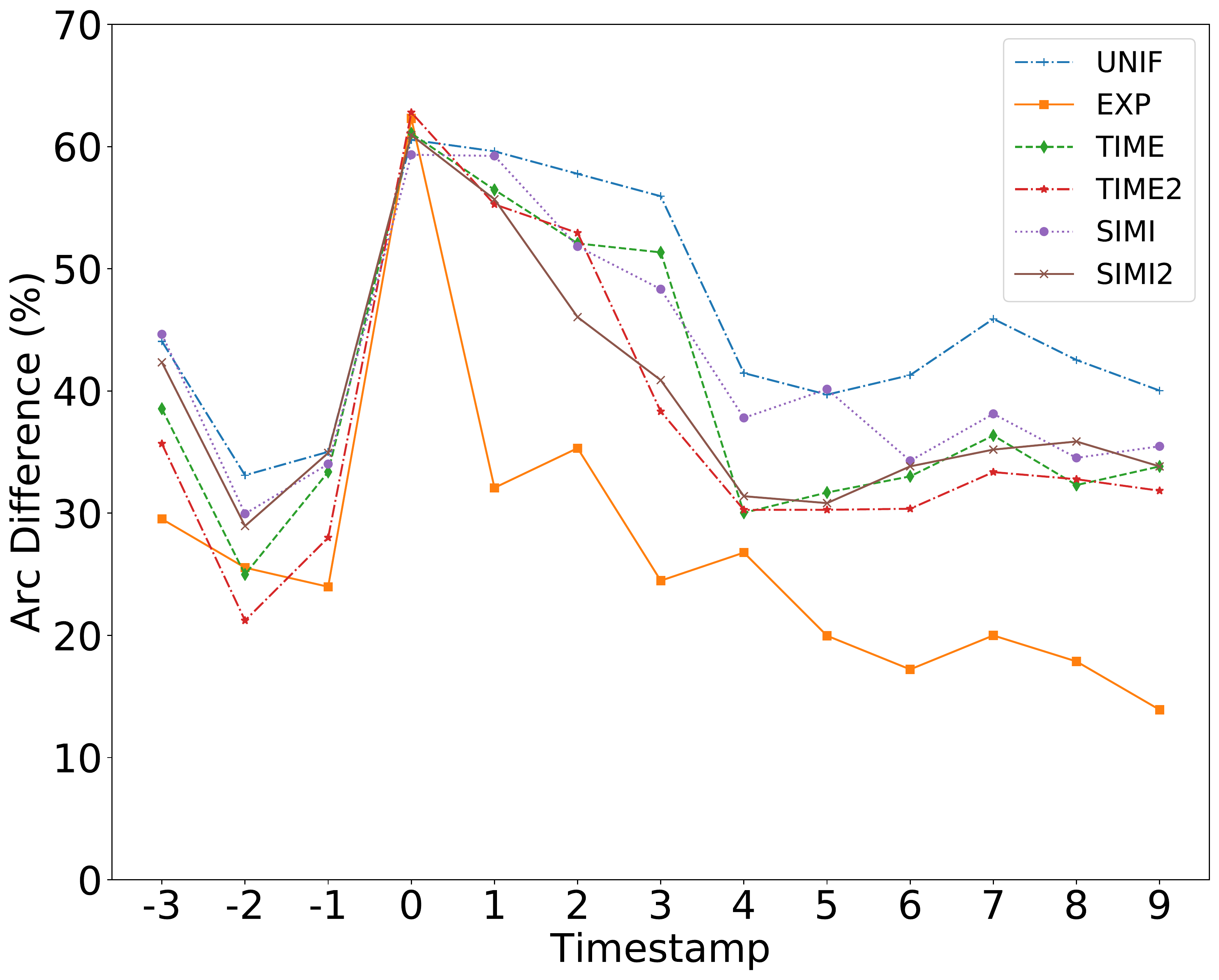}
    \end{minipage}
    \caption{Route and arc difference during concept drift (\underline{drop} in number of stops)}
    \label{fig:nonrev}
$ $

    \centering
    \begin{minipage}{0.4\textwidth}
        \centering
        \includegraphics[width=\linewidth]{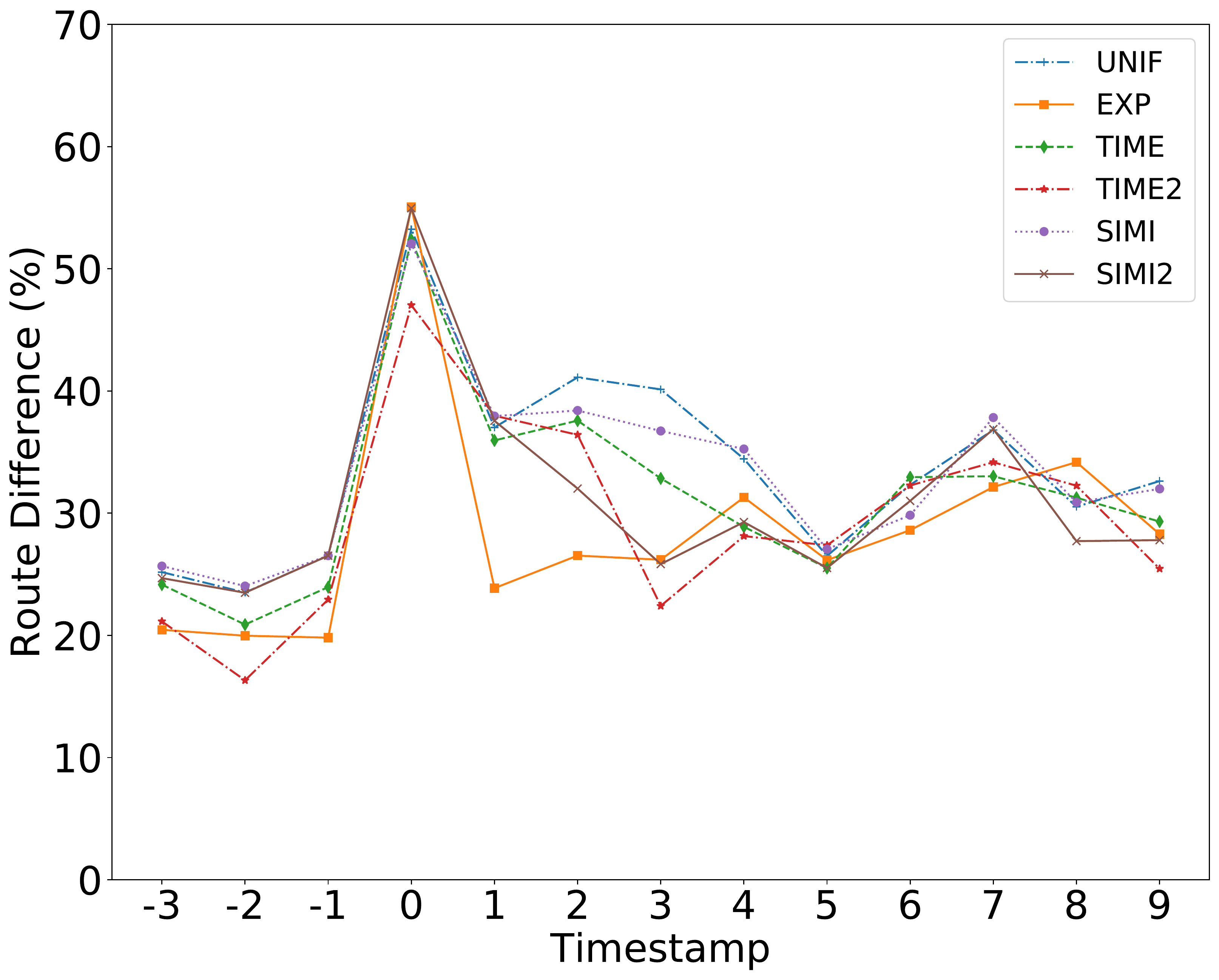}
    \end{minipage}%
    \begin{minipage}{0.08\textwidth}
        \hspace{0.1cm}
    \end{minipage}%
    \begin{minipage}{0.4\textwidth}
        \centering
        \includegraphics[width=\linewidth]{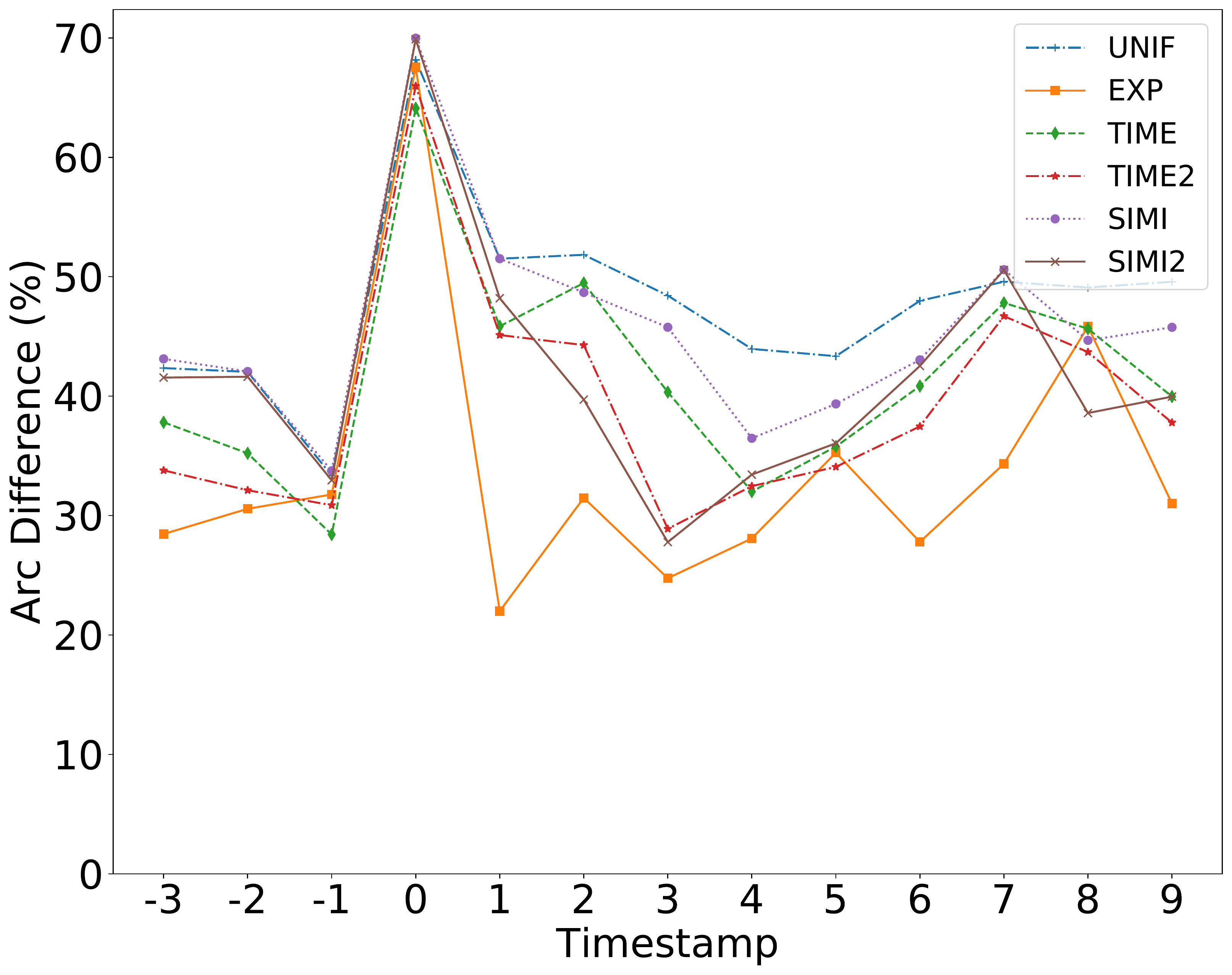}
    \end{minipage}
    \caption{Route and arc difference during concept drift (\underline{rise} in number of stops)}
    \label{fig:rev}
\end{figure}

We consider two scenarios. The first case is when there is an abrupt drop in the number of stops. Observe that this is the case immediately after week 53 in \textbf{Fig.~\ref{fig:visualization}}. For this scenario, we trim our data set such that our 13 test instances, i.e., instances in the \emph{tail} end of the data set, consist of 3 instances before the drift and 10 instances after the drift. As before, the probability matrix is trained on all data older than the ones contained in the test set. We plot the prediction accuracy of each scheme on each of the test instances.

The results of the first scenario are shown in \textbf{Fig.~\ref{fig:nonrev}}, where instance $0$ denotes the first instance after the drift. Naturally, in both route and arc difference, we observe a sharp rise in \emph{error} percentage at instance $0$. Especially with route difference, the error remained high for all the schemes (except EXP) until about instance $2$ or $3$, after which we see some improvement particularly for TIME2 and SIMI2. EXP, on the other hand, readjusted immediately after instance $0$ and clearly outperformed the other schemes in terms of prediction accuracy and its ability to adjust to changes in data structure. 

For the second scenario, we consider the case where there is a sudden rise in the number of stops. In order to use the same company data that we have, we simulate this case by reversing the time ranking of all the data instances. With the data in reverse order, we are able to simulate a drift where there is an increase in the number of stops, e.g., new stops that have not been seen before. As with the first case, we trim the data and select the 13 test instances \emph{after} reordering the data. Compared to the first case, here all the schemes seemed to \emph{adjust} more rapidly (\textbf{Fig.~\ref{fig:rev}}) after instance $0.$ The increase in route and arc difference at instance $0$ is unsurprisingly greater than in the previous scenario, as the schemes adjust with the new stops that were not seen before.

The above experiments show that generally, all schemes manage to stabilize after the initial rise in error due to structural change. In both scenarios, TIME2 and SIMI2 restabilize faster than their counterparts, TIME and SIMI. Among all the schemes, UNIF clearly performed the worst, while EXP was the fastest to adjust.  

\subsection{Addition of distance-based probabilities}
Up to this point, we have investigated the performance of our model using transition matrices made of only probabilities learned from the historical solutions. We expect, however, to gain further improvement when the learned transition probability matrix is combined with a distance-based probability matrix, the construction of which is described in Section~\ref{distancebasedprob}. For the next experiment, we investigate how the final solution is affected by the different ways of mixing, i.e., varying values of the $\beta$ parameter in Eq.~\eqref{eq:combinig_dis}.

\begin{figure}[t]

    \centering
    \begin{minipage}{0.45\textwidth}
        \centering
        \includegraphics[width=\linewidth]{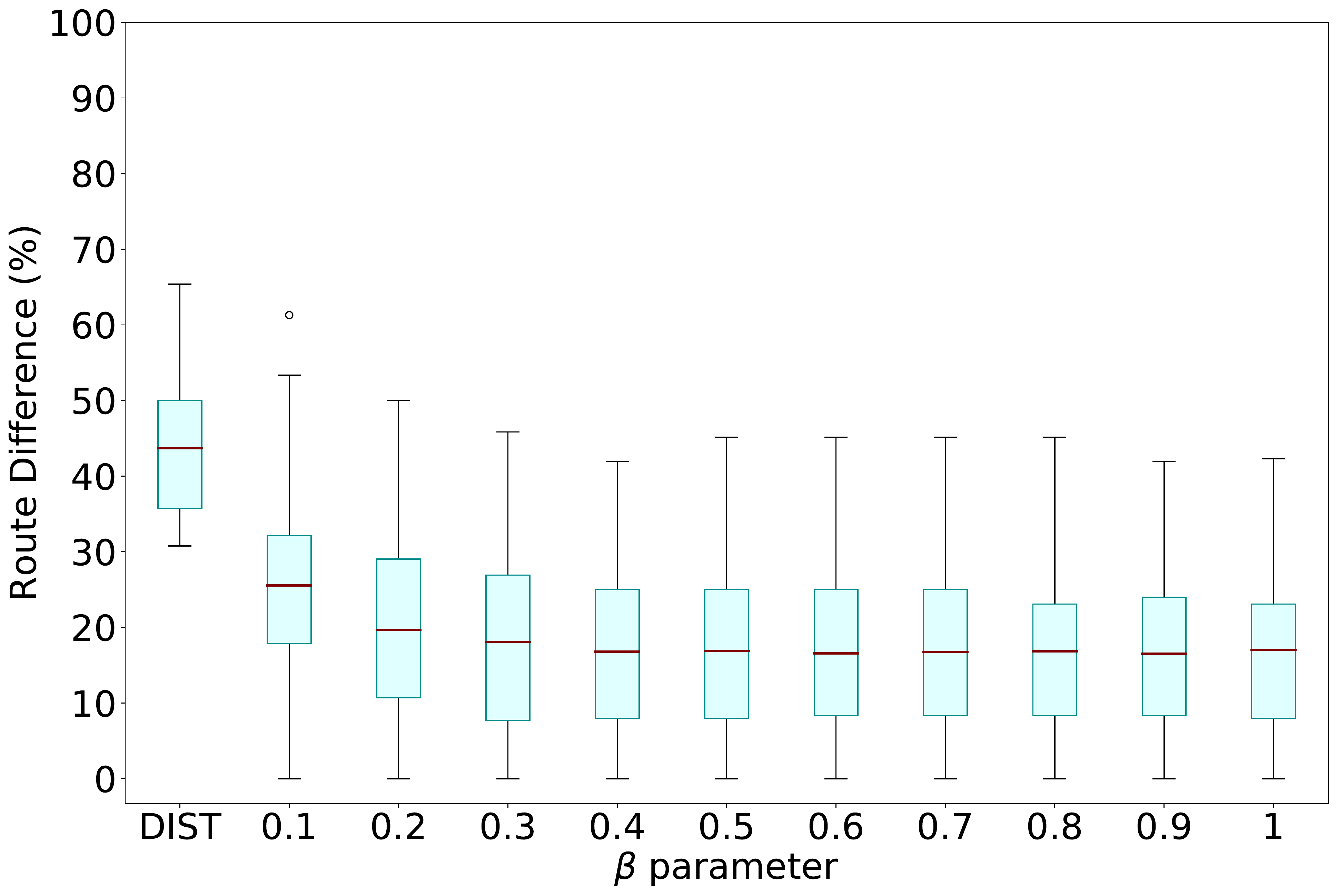}
    \end{minipage}%
    \begin{minipage}{0.08\textwidth}
        \hspace{0.1cm}
    \end{minipage}%
    \begin{minipage}{0.45\textwidth}
        \centering
        \includegraphics[width=\linewidth]{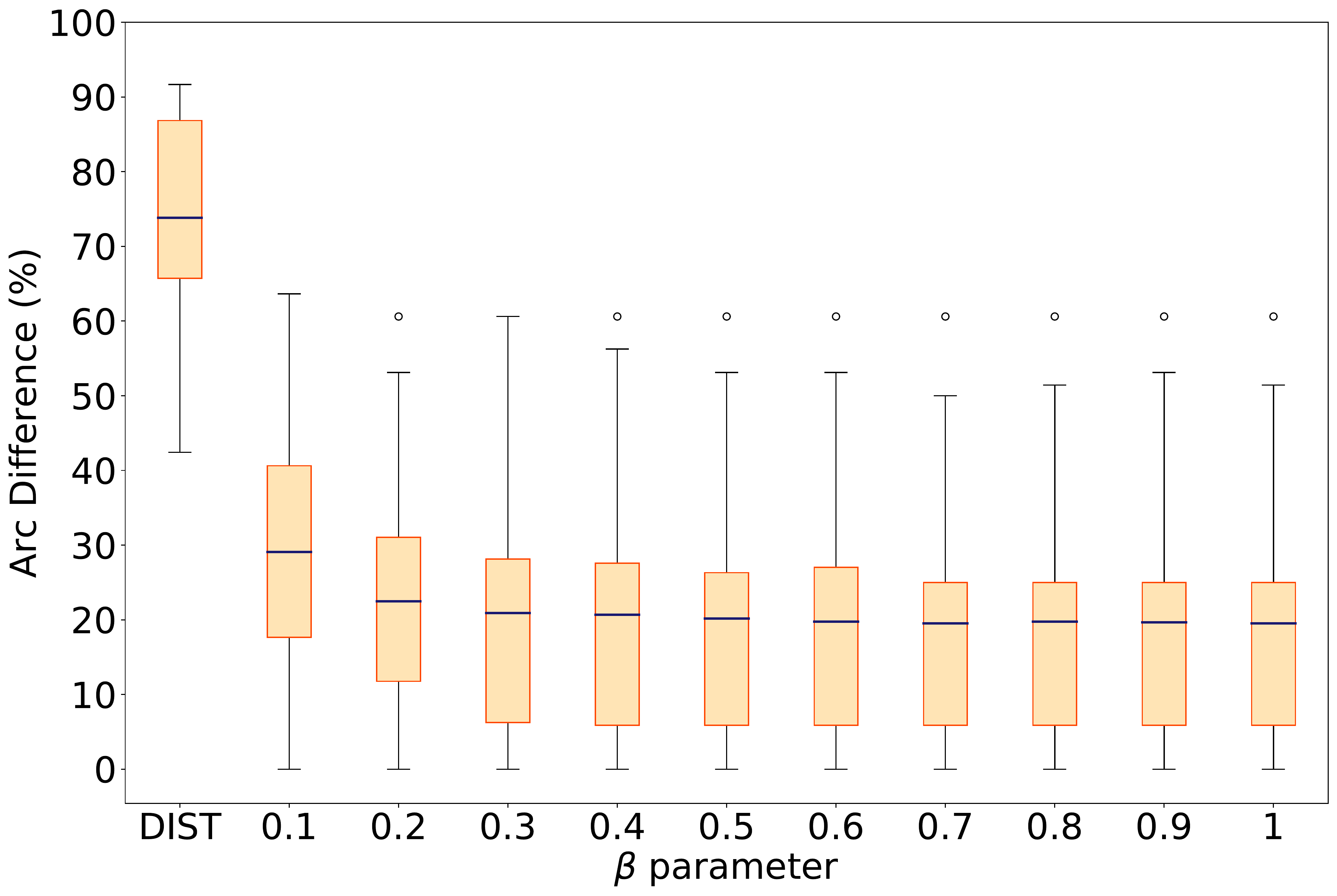}
    \end{minipage}

$ $

$ $

\begin{scriptsize}

{\tabcolsep=3pt\def\arraystretch{1.75}
\begin{tabular}{@{}lcccccccccccc@{}}
\toprule
 & \multicolumn{1}{l}{} & \multicolumn{10}{c}{$\beta$ parameter} & \multicolumn{1}{l}{} \\ \cmidrule(lr){3-12}
 & (DIST) & 0.1 & 0.2 & 0.3 & 0.4 & 0.5 & 0.6 & 0.7 & 0.8 & 0.9 & 1 & (Actual Sol) \\ \hline
RD & 43.70 & 25.56 & 19.68 & 18.09 & 16.78 & 16.90 & 16.57 & 16.73 & 16.85 & 16.54 & 17.02 & 0 \\
AD & 73.81 & 29.09 & 22.46 & 20.89 & 20.66 & 20.15 & 19.77 & 19.53 & 19.75 & 19.65 & 19.51 & 0 \\
Avg Total Dist (km) & 395.42 & 401.78 & 412.86 & 415.06 & 415.69 & 416.34 & 418.39 & 419.15 & 418.94 & 419.27 & 421.01 & 412.13 \\
Avg Time (s) & 5958.66 & 6.36 & 3.48 & 2.57 & 1.56 & 0.6004 & 0.5061 & 0.4265 & 0.5093 & 0.6751 & 0.7432 & - \\ \bottomrule
\end{tabular}}

$ $

\caption{EXP results for varying values of $\beta$ (entire period) \label{fig:beta} }
\end{scriptsize}

\end{figure}
%
\textbf{Fig.~\ref{fig:beta}} shows the result for different $\beta$ values, using EXP with the default scheme value on data from the entire period. 
When combined with the learned probability matrix, we see that even with small values of $\beta$, we already get better results than using distances alone.

While the boxplots alone seem to indicate that larger $\beta$ values lead to small but consistent improvements in RD and AD, we must warn that the arc and route differences with respect to the actual solution tell only one side of the story. In the accompanying table below the figures, we show the average RD/AD as well as the average total distance driven. 
Looking closely at the kilometers, we see that while higher values of $\beta$ generally lead to lower percent differences, they also increase the number of kilometers and do so beyond the number of kilometers of the actual solution. Using a $\beta$ value of 0.2 shows that on average, this leads to a good RD/AD accuracy as well as an average number of kilometers roughly equal to that of the actual solutions. We hence recommend a modest $\beta$ value, such as $\beta=0.2$, i.e., a mix of 20\%-preference and 80\%-distance probabilities.

\begin{figure}[htbp]
    \centering
    \begin{minipage}{0.295\textwidth}
        \centering
        \includegraphics[width=\linewidth]{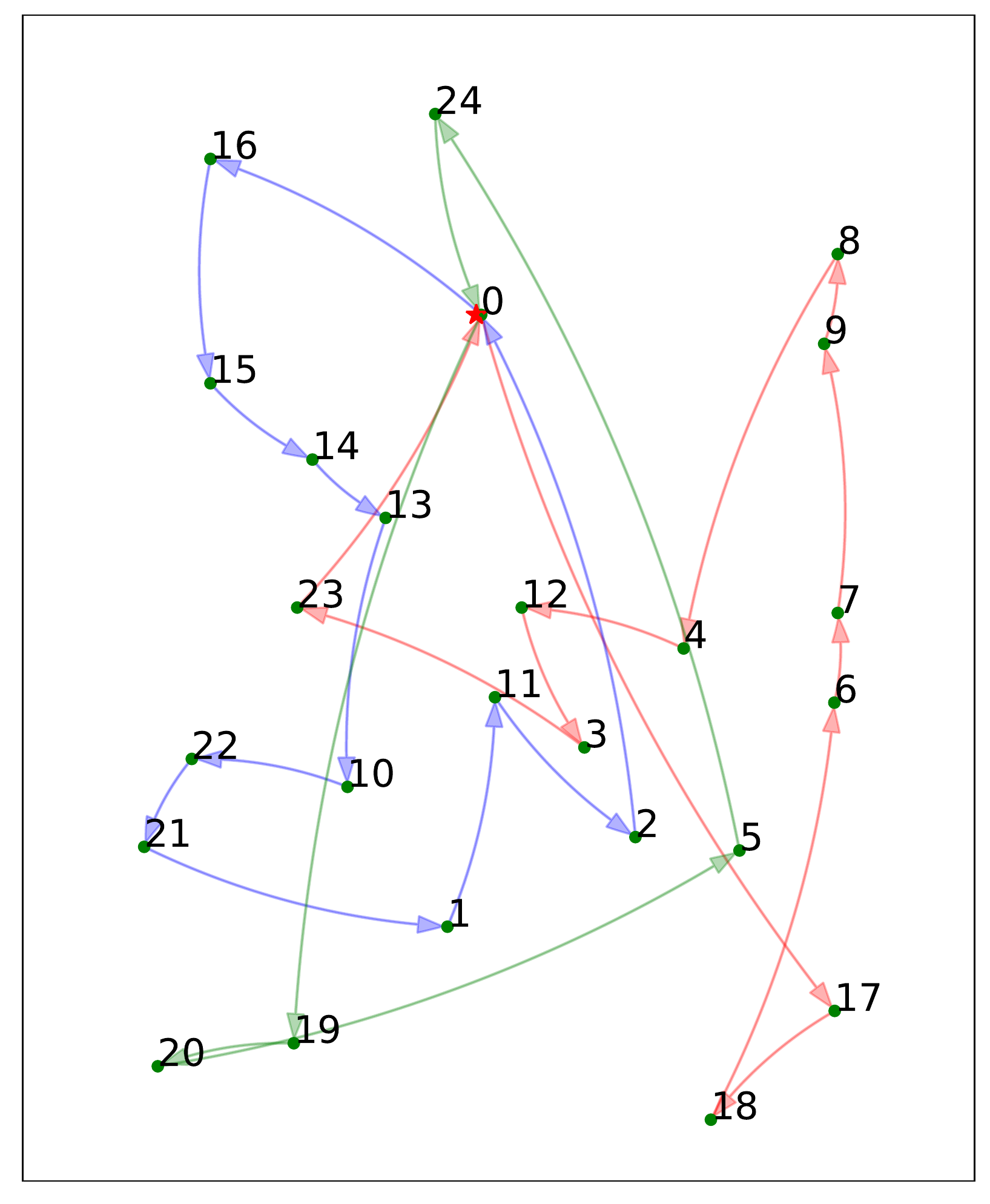}
        \caption{Distance-based solution \\ \phantom{}}
        \label{fig:map_dist}
    \end{minipage}%
    \begin{minipage}{0.01\textwidth}
        \hspace{0.1cm}
    \end{minipage}%
    \begin{minipage}{0.295\textwidth}
        \centering
        \includegraphics[width=\linewidth]{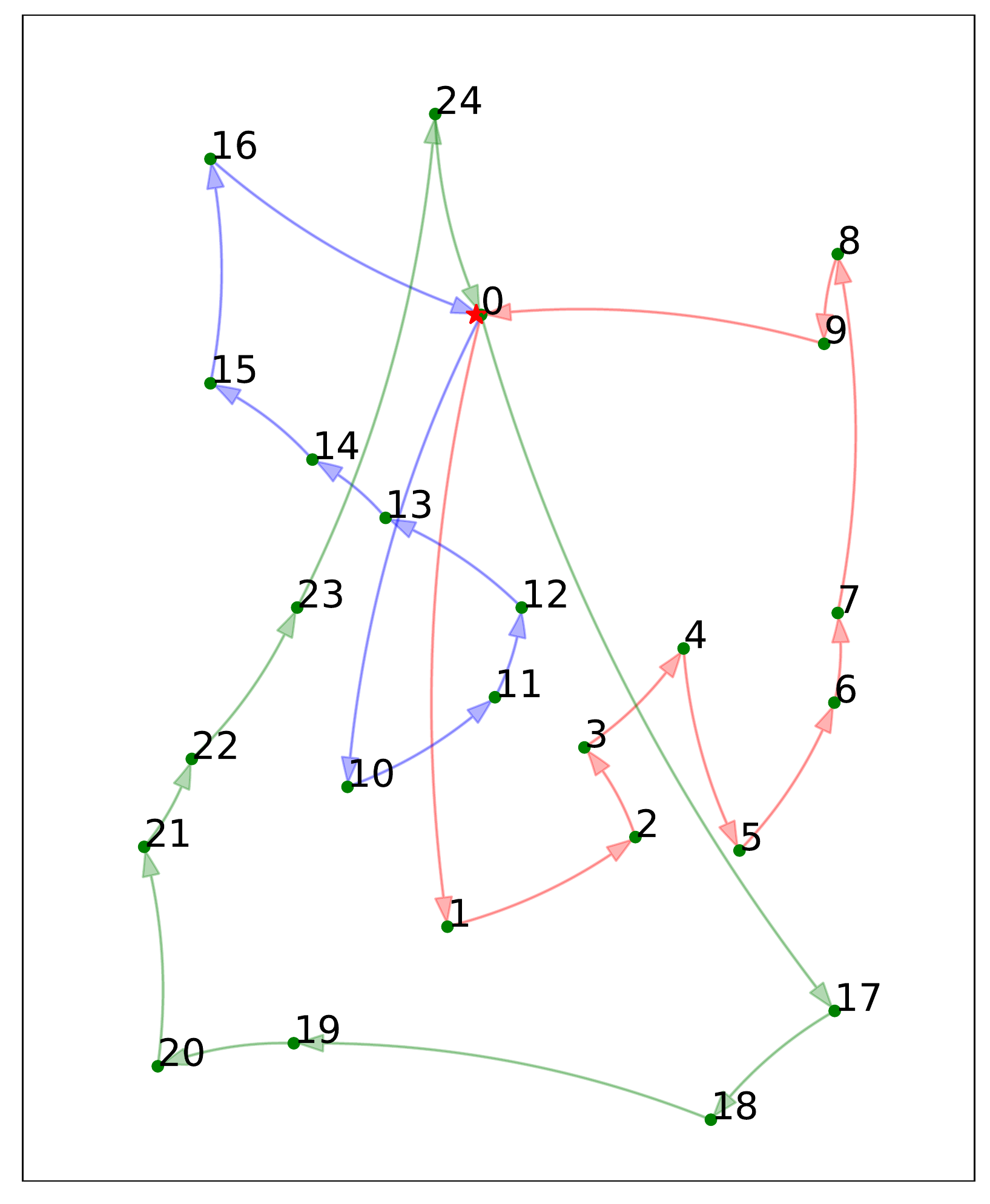}
        \caption{Human-made solution \\ \phantom{}}
        \label{fig:map_sol}
    \end{minipage}
    \begin{minipage}{0.01\textwidth}
        \hspace{0.1cm}
    \end{minipage}%
    \begin{minipage}{0.3155\textwidth}
        \centering
        \includegraphics[width=\linewidth]{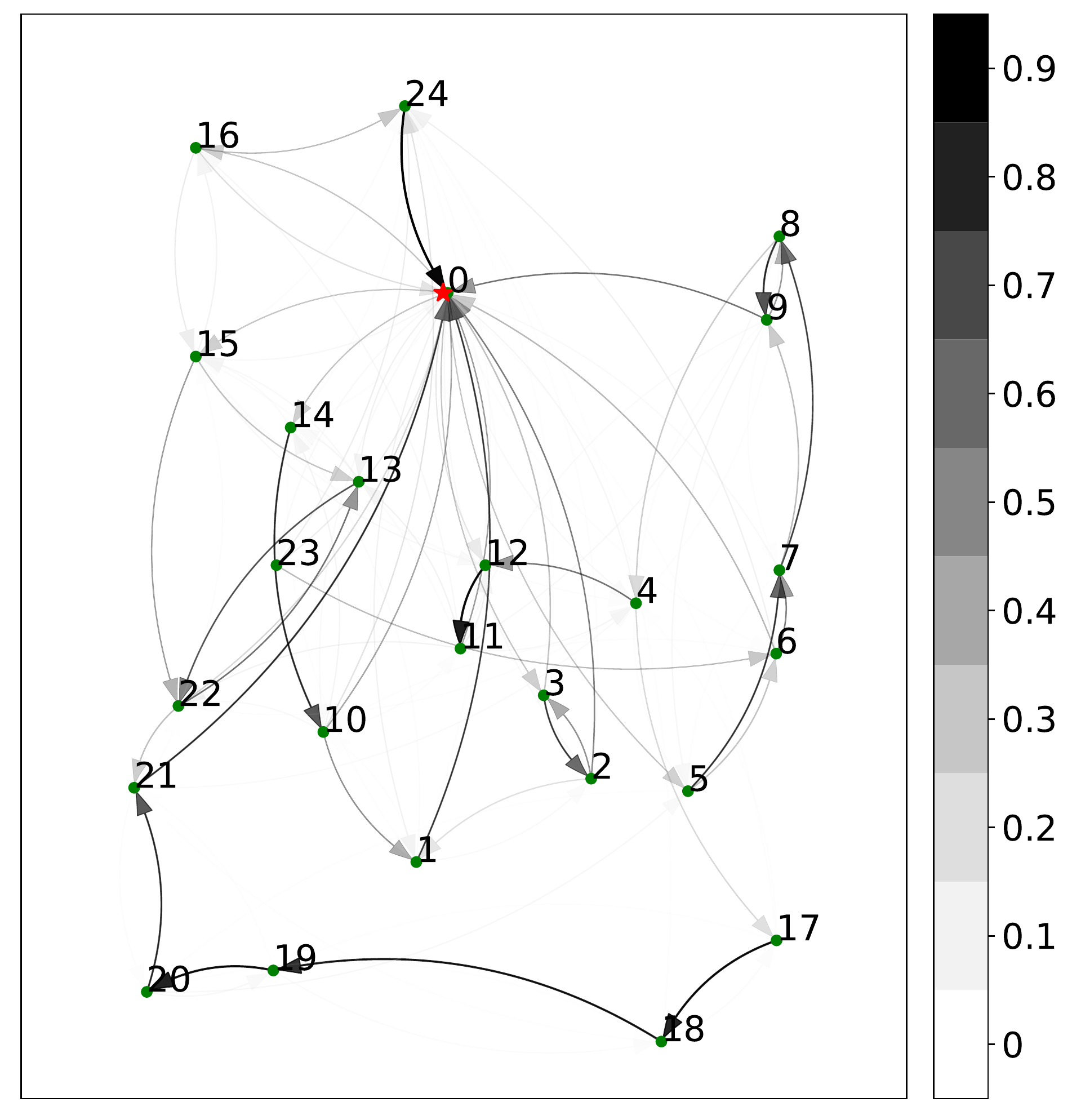}
        \caption{Learned first order\\ probabilities}
        \label{fig:map_trans}
    \end{minipage}

$ $

$ $

    \centering
    \begin{minipage}{0.305\textwidth}
        \centering
        \includegraphics[width=\linewidth]{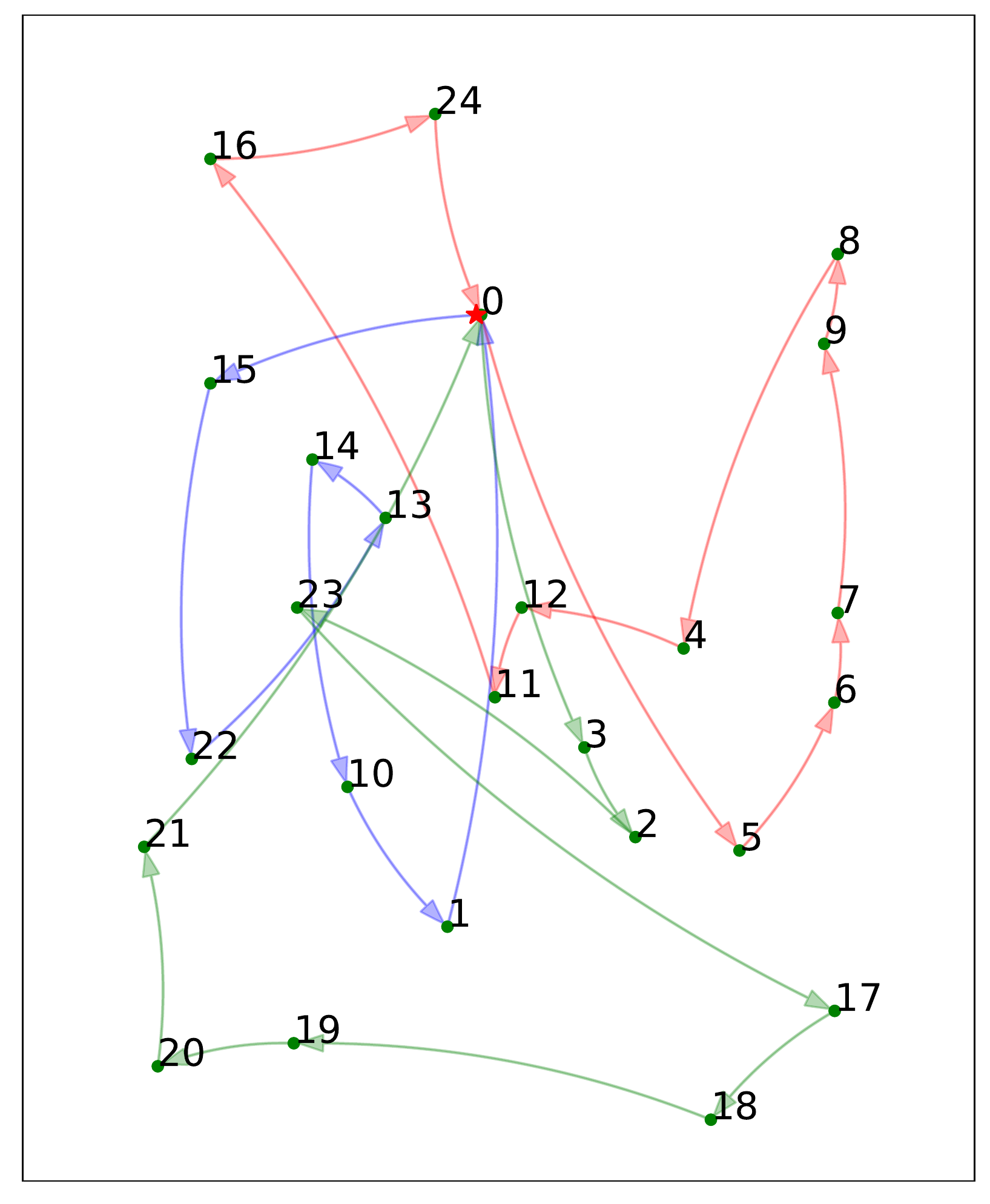}
        \caption{First order\\ solution}
        \label{fig:map_pred}
    \end{minipage}%
    \begin{minipage}{0.01\textwidth}
        \hspace{0.2cm}
    \end{minipage}%
    \begin{minipage}{0.305\textwidth}
        \centering
        \includegraphics[width=\linewidth]{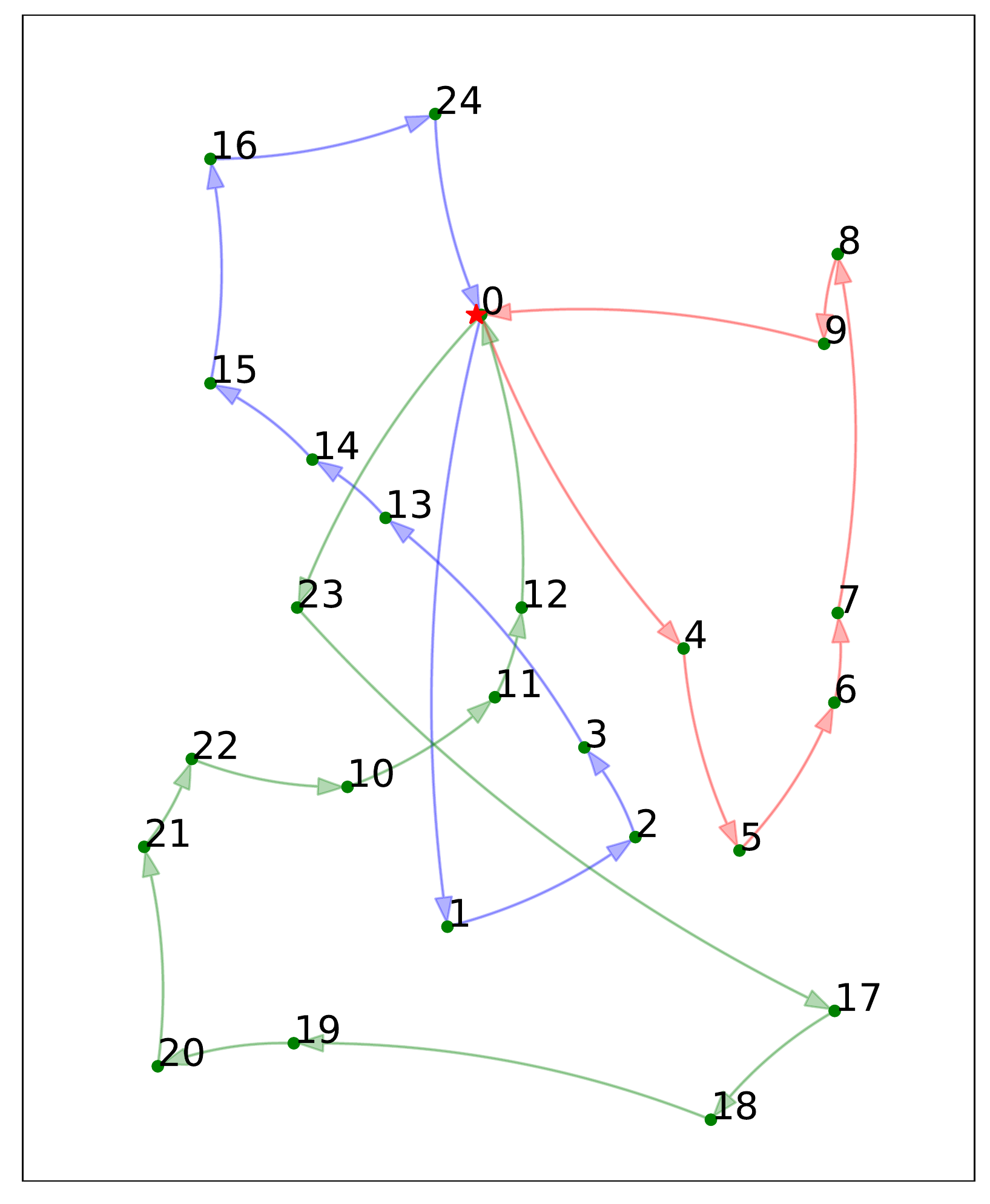}
        \caption{First order solution\\ with $\beta = 0.2$ }
        \label{fig:map_pred_beta}        
    \end{minipage}%
    \begin{minipage}{0.01\textwidth}
        \hspace{0.4cm}
    \end{minipage}%
    \begin{minipage}{0.305\textwidth}
        \centering
        \includegraphics[width=\linewidth]{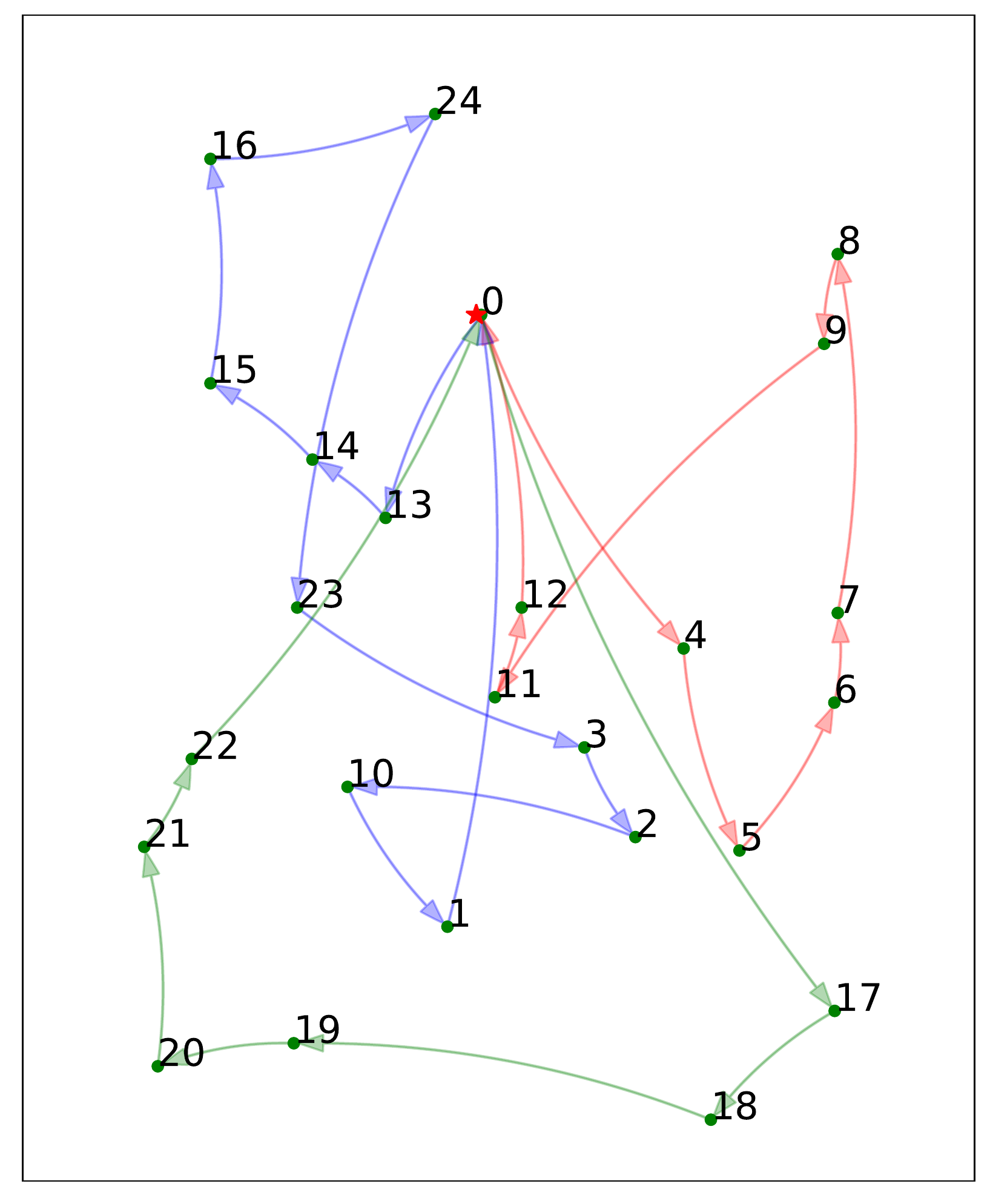}
        \caption{Second order\\ solution}
        \label{fig:map_pred_2mm}        
    \end{minipage}

$ $

$ $

\begin{minipage}{0.65\textwidth}
\centering
\begin{scriptsize}
{\tabcolsep=3pt\def\arraystretch{1.5}
\begin{tabular}{@{}lccccc@{}}
\toprule
 & DIST & First Order & \begin{tabular}[c]{@{}c@{}}First Order\\  ($\beta$ = 0.2)\end{tabular} & Second Order & \begin{tabular}[c]{@{}c@{}}Actual\\ (Human-made)\end{tabular} \\ \midrule
RD & 36.00 & 32.00 & 28.00 & 28.00 & 0 \\
AD & 81.48 & 70.37 & 25.93 & 72.63 & 0 \\
Total Dist (km) & 337.12 & 352.71 & 346.02 & 359.17 & 353.32 \\
Runtime (s) & 9938.91 & 0.2642 & 0.2580 & 21606.17 & - \\ \bottomrule
\end{tabular}}

$ $

\caption{Detailed results for example routing} \label{fig:detailed_results_ex}
\end{scriptsize}    
\end{minipage}     
\end{figure}

\subsection{Detailed Example}

A visual illustration of the way the routes are predicted by the transition matrix can be observed in \textbf{Figs. \ref{fig:map_dist}} - \textbf{\ref{fig:map_pred_2mm}}. \textbf{Fig.~\ref{fig:map_dist}} shows a map (figure not drawn to scale) of 24 stops to visit using three delivery vehicles. The depot is denoted by 0. Also shown in \textbf{Fig.~\ref{fig:map_dist}} is the solution obtained by using the standard distance-based approach. The actual routing used by the company is shown in \textbf{Fig.~\ref{fig:map_sol}}. Note that except for the slight similarity in the way the stops are clustered into routes, the two routings appear to be almost completely different. 

From the historical data of the same weekday as the actual solution, we construct a transition probability matrix. The first order matrix learned with the simplest scheme (UNIF) 
is visualized in \textbf{Fig. \ref{fig:map_trans}}, where darker arcs indicate higher probabilities. It can be observed that the image shows a clear structure, with distinct connections, e.g., to the furthest stops, but also a higher variability in the denser regions and near the depot. \textbf{Fig. \ref{fig:map_pred}} shows the first order solution constructed with the probability matrix of \textbf{Fig. \ref{fig:map_trans}}. We can see that the solution was able to capture key structural parts while making trade-offs elsewhere to come up with a global solution, e.g., making a connection from stop 11 to 16. The actual human-made solution, on the other hand, was made with a number of distinct choices which cannot be easily predicted just by looking at the probability matrix map, e.g., the direct connections from 0 to 17, and 23 to 24. 

The solution obtained by adding distance-based probabilities to the first order model is displayed in \textbf{Fig. \ref{fig:map_pred_beta}}. It can be observed that the solution is able to keep the general structure of the human-made solution, while also eliminating the longer connections (e.g., 11 to 16, and 15 to 22) made by the purely preference-based first order model.

\textbf{Fig. \ref{fig:map_pred_2mm}} shows the solution when optimising with the second order model,
which improved the route difference of the first order solution (RD  in \textbf{Fig. \ref{fig:detailed_results_ex}}). Also from the detailed table of results, we can see that in this particular example, the first order with $\beta$ model was able to most closely capture the general structure of the human-made solution. Obviously, this is not always the case (see \textbf{Fig. \ref{fig:beta}} in Section~\ref{drift_expt}). Nevertheless, looking at the total distances displayed in \textbf{Fig. \ref{fig:detailed_results_ex}}, here we observe that adding distance-based probabilities not only preserved the structure but also offered a solution with fewer kilometers. Hence, it does not merely mimic the human planners but is also able to take preferences into account \emph{while} still optimizing the total distance.  


\section{Concluding Remarks} \label{s:conclusions}
One of the crucial first steps in solving vehicle routing problems is explicitly formulating the problem objectives and constraints. 
Oftentimes in practice, the optimized routing takes into account not only time and distance-related factors, but also a multitude of other concerns, which remain a tacit knowledge of the route planners.   
Specifying each sub-objective and constraint may be an intractable task. Moreover, as we have observed in practice, computed solutions seldom fully satisfy the wishes of the route planners and all involved stakeholders. 

This paper studied the potential of learning the preferences of the route planners from historical solutions in solving VRP.
Specifically, we presented an approach to solving the VRP which does not require a full explicit problem characterization. Inspired by existing research on exploiting Markov models to predict individual route choices, we developed an approach that learns a probabilistic preference model based on historical solutions. This probabilistic model can subsequently be optimized to obtain the most \textit{likely} routing for an entire fleet. 
We have shown how this approach is capable of learning implicit preferences from real data, resulting in more accurate solutions than  using distances alone. The algorithm performs well even without capacity demands, confirming its ability to learn patterns in the underlying solution structure.

Our work extends beyond a first order Markov model and we have provided a more general formalization, which includes a second order Markov model, for learning and optimizing preferences.
We have seen that the second order model improves the accuracy in terms route difference. This improvement, however, does not translate to arc difference. Furthermore, the second order model is expensive in terms of computational burden. Here, we assert that a richer database is necessary to better evaluate the performance of higher order models.

We also presented an approach which combines distance-based and preference-based probabilities. This approach has the advantage of being robust to changing client sets and computationally fast due to sparsity of the resulting cost matrix. By mixing the two probabilities, we obtained more robust results without sacrificing computational times. 

 Results on real data have been encouraging. Validation on other real-life data, however, should be considered for generalization, as other data may have more (or less) structural assumptions. Our proposed first order approach could be plugged into existing VRP software today, but as with all predictive techniques there should be human oversight to avoid unwanted bias.

Future work on the routing side will involve applications to richer VRP model, e.g., problems involving time windows, multiple deliveries, etc. On the learning side, the use of richer predictive models such as neural networks or higher order Markov models can be considered. 
Also, using separate learned or contextual models per vehicle or per driver is worth investigating. Finally, extending the technique so that the user can be actively queried and learned from during construction is an interesting direction, e.g., to further reduce the amount of user modifications needed on the predicted solutions.


\printbibliography

\begin{appendix}
\section{UNIF without and with capacity constraints}
\begin{figure}[htbp]
\centering
\begin{subfigure}{.4\textwidth}
  \centering
        \includegraphics[width=0.99\linewidth]{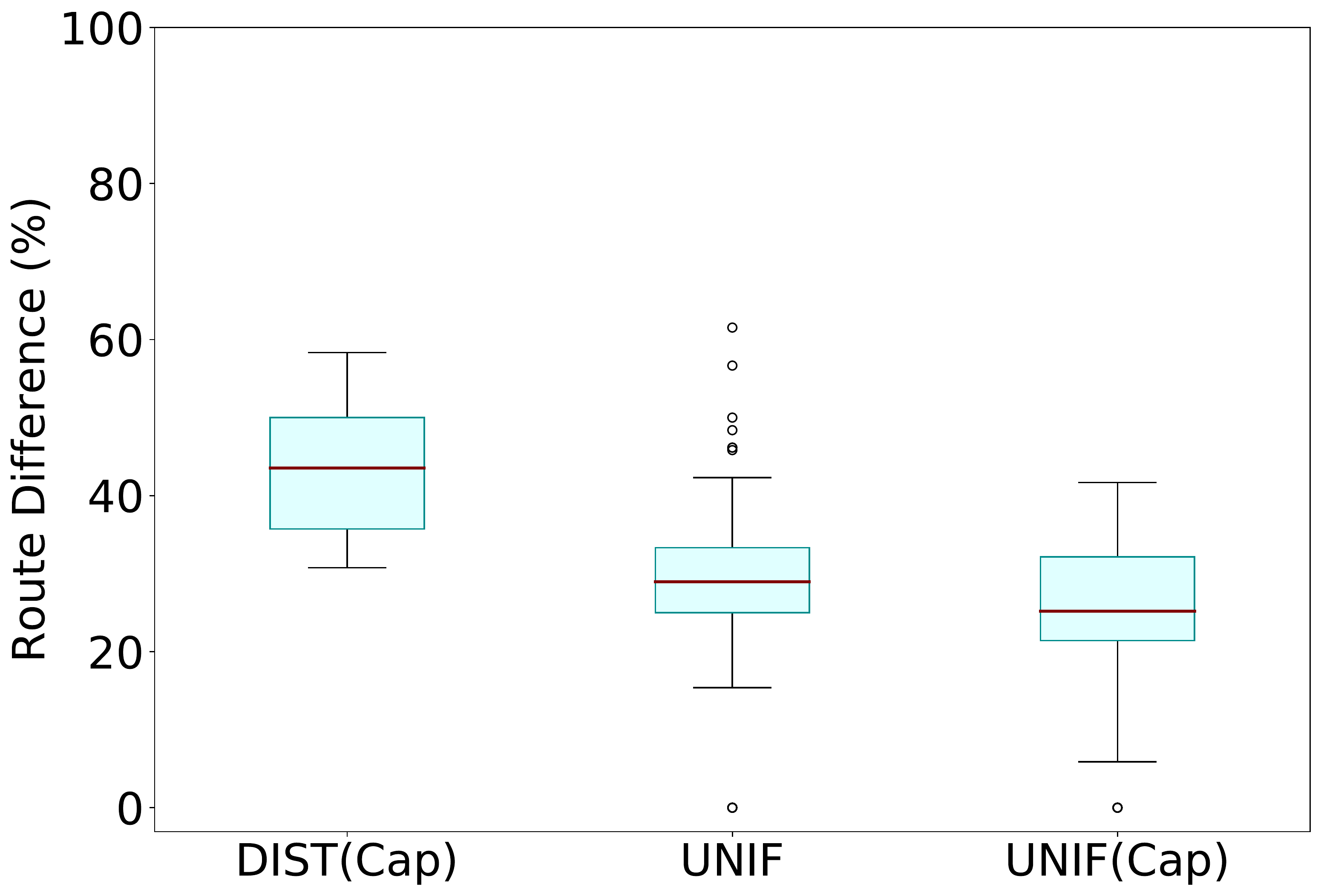}
\end{subfigure}
    \begin{minipage}{0.08\textwidth}
        \hspace{0.1cm}
    \end{minipage}
\begin{subfigure}{.4\textwidth}
\centering
        \includegraphics[width=\linewidth]{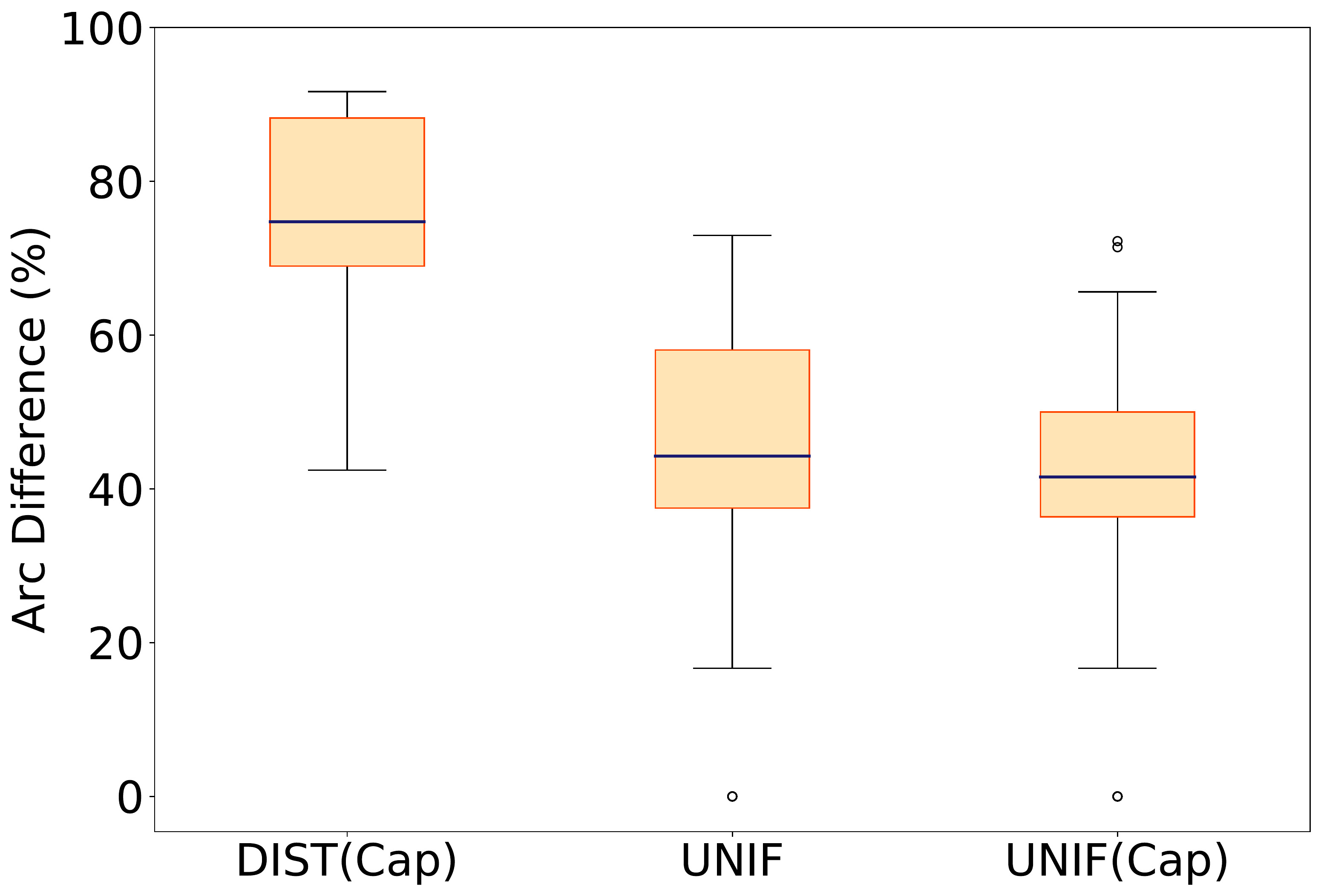}
\end{subfigure}

$ $

$ $

\begin{scriptsize}
{\tabcolsep=3pt\def\arraystretch{1.5}
\begin{tabular}{l|c|c|c}
Weighing Scheme & \multicolumn{1}{c|}{\ DIST(Cap) \quad } & \multicolumn{1}{c|}{\quad UNIF \ \; } & \multicolumn{1}{c}{\ UNIF(Cap)} \\ \hline
Avg Computation Time (s) & 5958.66 & 0.1135 & 0.9730 
\end{tabular}}

$ $

\caption{UNIF without and with capacity (Cap) constraints (data from entire period)}
\label{fig:WithCaps}
\end{scriptsize}

\end{figure}

This experiment was done to compare the prediction accuracy of UNIF without and with the capacity (Cap) demand estimates (\textbf{Fig. \ref{fig:WithCaps}}). The motivation is to investigate how UNIF will perform even without the capacity constraints. 
When evaluating without the capacity estimates, in order to keep the subtour elimination constraint (\ref{eqn:con_cap1}), each $q_i$ is assigned a value of $1$ while using the number of stops as a fictive bound on the vehicle capacities, e.g., $Q\!=\!n$.
As a baseline, we included the solution (DIST) obtained by solving the standard distance-based CVRP to near-optimality (5\% optimality gap). Computation was done on data from the entire period.

Results show that DIST is consistently outperformed by UNIF.
Moreover, in both route and arc difference, we always notice some improvement when capacity estimates are taken into account. Remarkably, when using the transition probability matrices, we see that we can solve the VRP even \textit{without} the capacity constraints and still get meaningful results. This shows the ability of the method to learn the structure underlying the problem just from the solutions.

As for the computation time, DIST needed an average of almost 6000 seconds to solve \emph{each} instance despite the imposed near-optimality relaxation. In contrast, both UNIF and UNIF(Cap) only took less than a second on average to obtain the optimal solutions, with UNIF(Cap) taking slightly more time due to the additional capacity constraints. Additionally, we observed that the learned probability matrices are much more sparse (containing more 0 or near-0 values) than the distance matrices.

\section{Parameter Sensitivity}


\begin{figure}[t]
    \centering
    \begin{minipage}{0.4\textwidth}
        \centering
        \includegraphics[width=\linewidth]{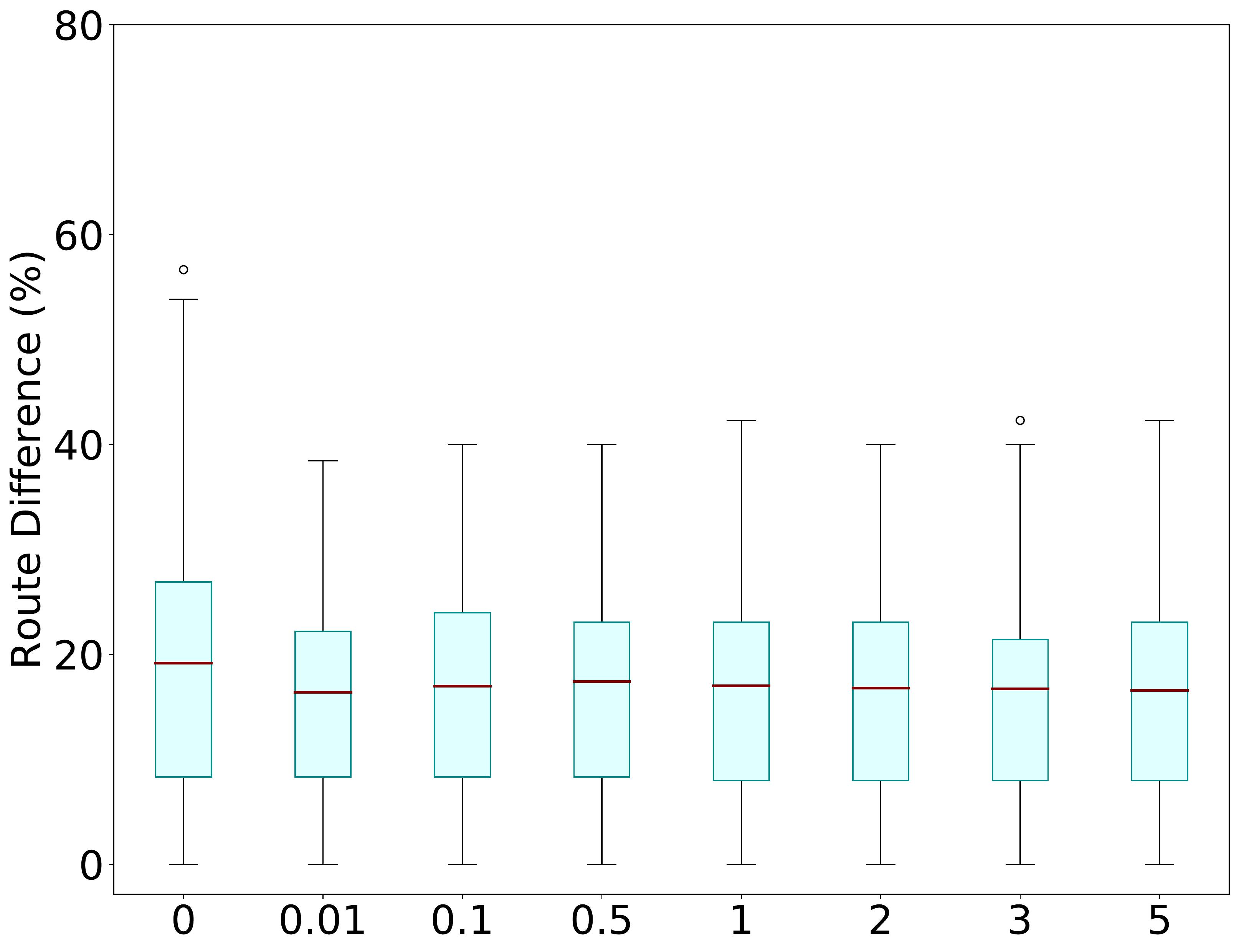}
    \end{minipage}%
    \begin{minipage}{0.08\textwidth}
        \hspace{0.1cm}
    \end{minipage}%
    \begin{minipage}{0.4\textwidth}
        \centering
        \includegraphics[width=\linewidth]{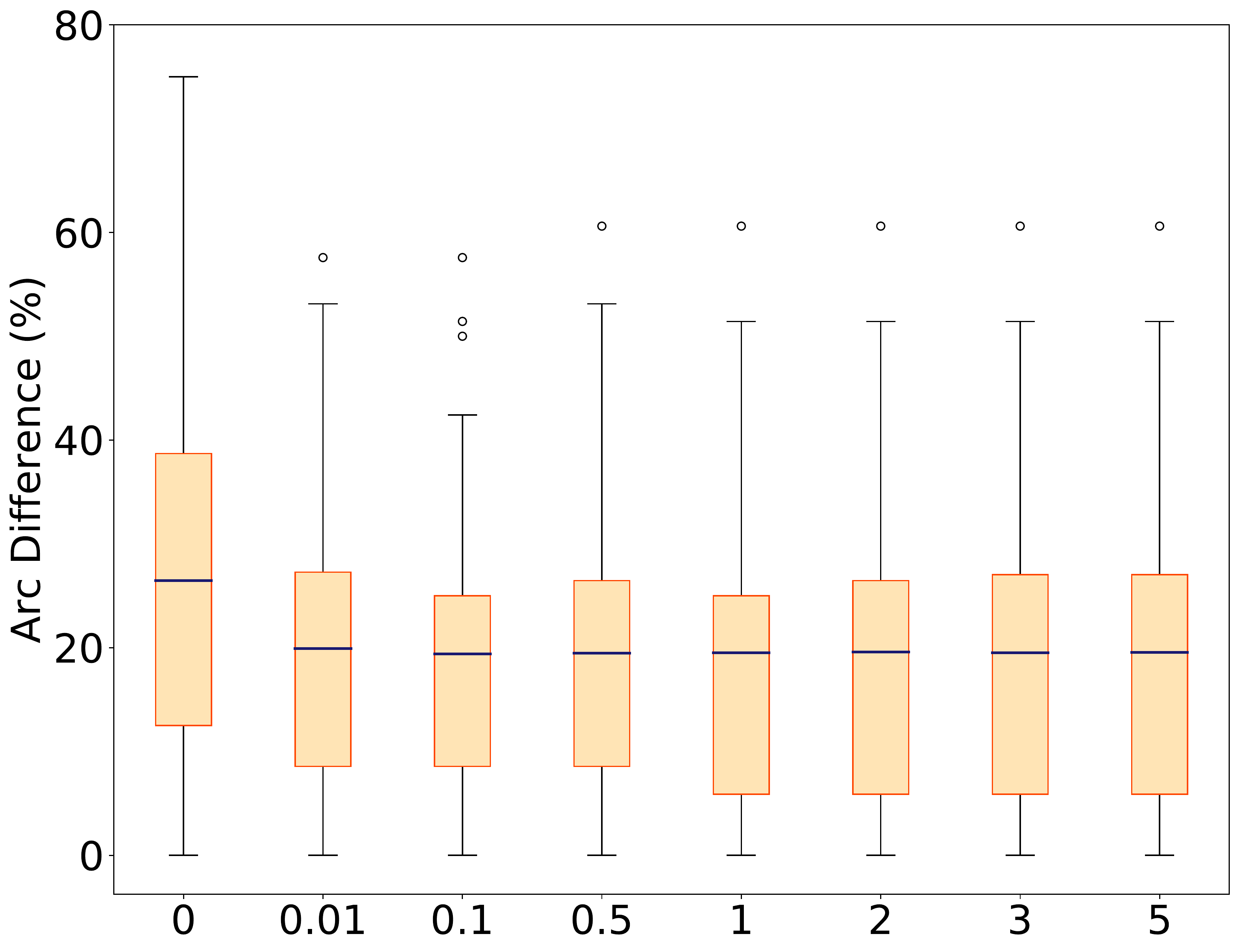}
    \end{minipage}
\caption{Route and arc difference for values of Laplace parameter $\lambda$ (entire period)}
\label{fig:Laplace}
$ $

    \centering
    \begin{minipage}{0.4\textwidth}
        \centering
        \includegraphics[width=\linewidth]{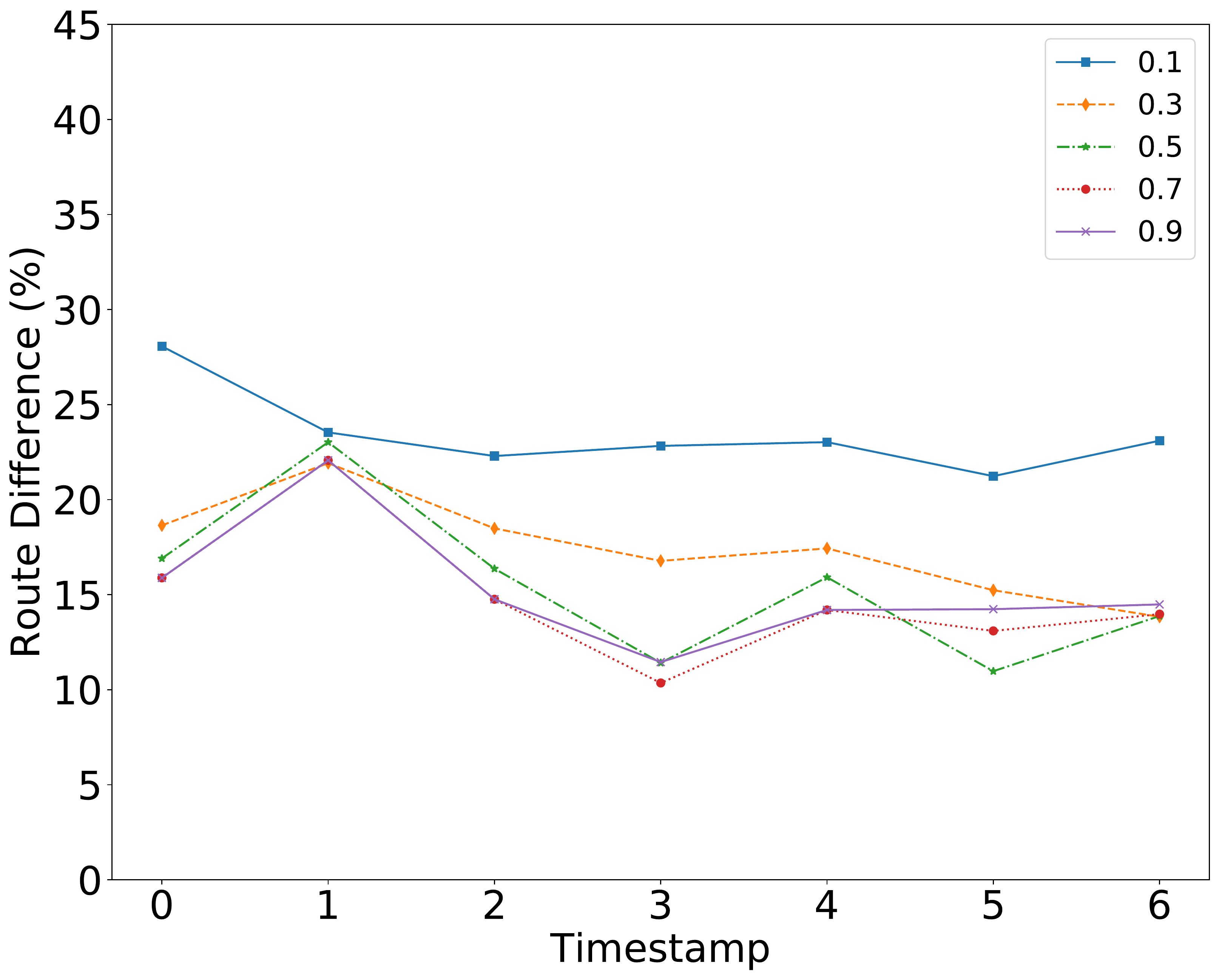}
    \end{minipage}%
    \begin{minipage}{0.08\textwidth}
        \hspace{0.2cm}
    \end{minipage}%
    \begin{minipage}{0.4\textwidth}
        \centering
        \includegraphics[width=\linewidth]{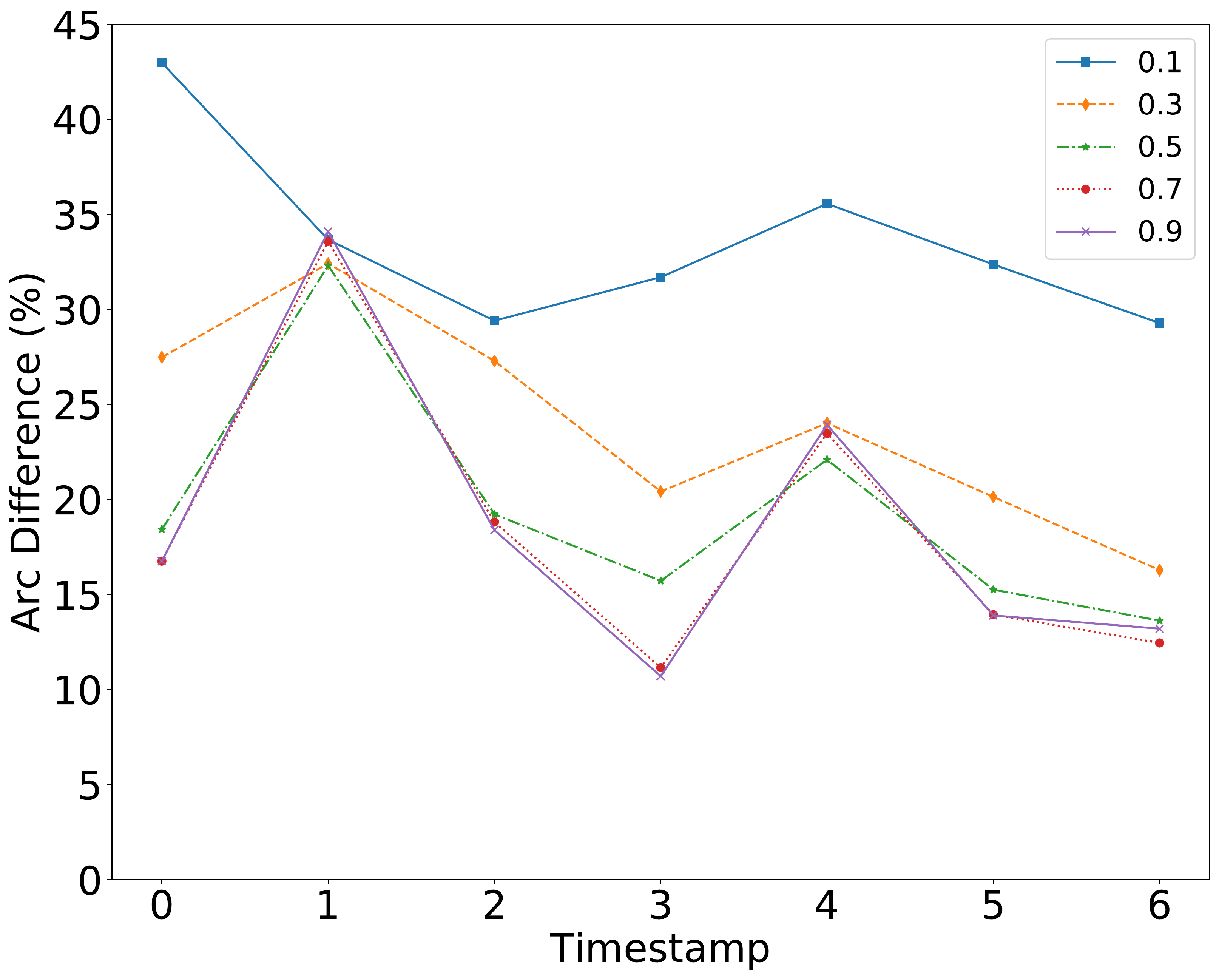}
    \end{minipage}
    \caption{Route and arc difference for varying values of EXP parameter (entire period)}
    \label{fig:ExpParameter}
\end{figure}

\noindent\textbf{Laplace parameter $\lambda$.}
To understand the effects of varying the Laplace parameter $\lambda$ (\textbf{Fig. \ref{fig:Laplace}}), we performed an experiment on data from the entire period. 
Capacity demands were taken into account and EXP was selected on the basis of the results from the previous experiments. It is interesting to note that EXP worked well even with no smoothing ($\lambda=0$). It can also be observed that EXP produced slightly improved results when smoothing is applied.
In general, however, we see that on our data, Laplace smoothing has very little effect.

\vspace{1em}
\noindent
\noindent\textbf{EXP parameter $\alpha$.}
We examined the behavior of the model for different values between 0 and 1 of the EXP parameter $\alpha.$ Instead of looking at the average percent differences as we did for the preceding experiment on the Laplace parameter, here we opted to inspect more closely the evolution of the prediction across different time periods. In \textbf{Fig.~\ref{fig:ExpParameter}}, each point on the graph represents the average route or arc difference when the model is tested on the test instances from the corresponding time period $0,1,\ldots,$ or $6$, with $0$ being the oldest, and $6$ the newest time period. The results of the experiment show that as $\alpha$ increases, prediction accuracy also increases. Accuracy appears to stabilize at $\alpha=0.7,$ which incidentally coincides with our choice of the parameter's default value.
\section{Distance-based Probabilities }\label{append:distance}

Here, we prove that minimizing the absolute distances in the standard CVRP is equivalent to minimizing the distance-based probabilities of Eq.~\eqref{eqn:invcost} under some mild conditions. We write the solution using the distance-based probabilities as:
\begin{align}
    & \argmin_{x} \sum_{(i,j) \in A} \hat{d}_{ij} x_{ij} \nonumber 
    = \argmin_{x} \sum_{(i,j) \in A} -\log\,\Big(\frac{e^{-d_{ij}}}{\sum_k e^{-d_{ik}}}\Big)x_{ij} \nonumber 
     = \argmin_{x} \sum_{(i,j) \in A} - \Bigg(  \log e^{-d_{ij}} - \log \sum_k e^{-d_{ik}} \Bigg)x_{ij} \nonumber\\
     =& \argmin_{x} \sum_{(i,j) \in A} d_{ij}x_{ij} + \sum_{i \in V} \sum_{j: (i,j) \in A} \log \bigg(\sum_k e^{-d_{ik}} \bigg) x_{ij}  \nonumber \\
     = & \argmin_{x} \sum_{(i,j) \in A} d_{ij}x_{ij} +  \sum_{i \in V\setminus \{0\}} \log \bigg(\sum_k e^{-d_{ik}} \bigg) \sum_{j: (i,j) \in A} x_{ij} +  \log \bigg(\sum_k e^{-d_{0k}} \bigg) \sum_{j: (0,j) \in A} x_{0j}. 
     \label{eq:dist_append}
\end{align}
The first term in the final expression corresponds to the classical VRP where the total distance is minimized. Given that $\sum_{j: (i,j) \in A} x_{ij}=1$ (Eq. \eqref{eqn:con_flow1}), the second term is always a constant. Hence, it does not play any role in the optimization above.  The third term depends on the number of vehicles used in the routing $\bf{x}$.

Whenever the optimal solution utilizes all the vehicles, we have $\sum_{j: (0,j) \in A} x_{0j } = m $ (where $m$ is the number of vehicles). 
As the third term also becomes a constant, Eq.~\eqref{eq:dist_append} simply becomes $\argmin_{x} \sum_{(i,j) \in A} d_{ij}x_{ij} $, which is the same as the original CVRP objective function. 

From the operational point of view, in the company setting in which this work is applied, the number of available drivers each day is fixed. Allocating a route to each of the drivers does not entail any additional cost. Therefore, we can assume that the equality constraint is always active.

Otherwise, we can always scale up the softmax function with a parameter $\theta >0$ in order to obtain $g(\theta)=\sum_k e^{-\theta d_{0k}} =1$. In fact, $g(\cdot)$ is a continuous function such that $g(0)=|V|$ 
and $\lim_{\theta\rightarrow +\infty} g(\theta) = 0$. Hence, there exists a value $\theta^* >0$ with the desired property. 
Thus, scaling up by $\theta^*$ makes the third term in Eq.~\eqref{eq:dist_append} equal to 0, and the solution using the distance-based probabilities is always equivalent to the CVRP solution.

 \end{appendix}

\end{document}